\documentclass[journal]{IEEEtran}
\IEEEoverridecommandlockouts
\usepackage{amsmath,amssymb,amsthm}
\usepackage{graphicx}
\usepackage{bmpsize}
\usepackage{tabularx}
\usepackage{hyperref}
\usepackage{blindtext}
\usepackage[belowskip=-7pt,aboveskip=-2pt]{caption}
\usepackage{textcomp}
\usepackage{multicol}
\usepackage{caption}
\usepackage{subcaption}
\usepackage{mathtools}
\usepackage{float}
\usepackage[linesnumbered,algoruled]{algorithm2e}

\SetKwComment{Comment}{$\triangleright$\ }{}
\SetCommentSty{itshape} 
\SetAlFnt{\small}
\SetKwComment{tcp}{\small // }{}%
\SetCommentSty{small}
\newcommand\tab[1][0.3cm]{\hspace*{#1}}
\newcommand\stab[1][0.7cm]{\hspace*{#1}}
\newcommand\mtab[1][1.1cm]{\hspace*{#1}}

\begin{document}

\title{A Fast Edge-Based Synchronizer for Tasks in Real-Time Artificial Intelligence Applications
}
\author{Richard~Olaniyan and~Muthucumaru~Maheswaran
\thanks{This research is supported by an Ericsson/Mitacs Accelerate grant, NSERC Discovery grant and PTDF Nigeria/PRESSID funding.}
\thanks{R. Olaniyan is with the School of Computer Science, McGill University, Montreal,
QC, Canada (e-mail: richard.olaniyan@mail.mcgill.ca).}
\thanks{M. Maheswaran is with the Department of Electrical and Computer Engineering, McGill University, Montreal,
QC, Canada (e-mail: muthucumaru.maheswaran@mcgill.ca).}

}

\maketitle

\begin{abstract}
Real-time artificial intelligence (AI) applications mapped onto edge computing
need to perform data capture, process data, and device actuation within given
bounds while using the available devices. Task synchronization across the
devices is an important problem that affects the timely progress of an AI
application by determining the quality of the captured data, time to process the
data, and the quality of actuation. In this paper, we develop a fast edge-based
synchronization scheme that can time align the execution of input-output tasks
as well compute tasks. The primary idea of the fast synchronizer is to cluster
the devices into groups that are highly synchronized in their task executions
and statically determine few synchronization points using a game-theoretic
solver. The cluster of devices use a late notification protocol to select the
best point among the pre-computed synchronization points to reach a time aligned
task execution as quickly as possible. We evaluate the performance of our
synchronization scheme using trace-driven simulations and we compare the
performance with existing distributed synchronization schemes for real-time AI
application tasks. We implement our synchronization scheme and compare its training accuracy and training time with other parameter server synchronization frameworks.\\


\end{abstract}

\section{Introduction}

Intelligent systems such as self-driving cars and robots can work either autonomously (i.e., doing all necessary processing by themselves) or work collaboratively with other self-driving cars or robots and environmental infrastructures~\cite{shin2017intelligent}. For instance, self-driving cars can collaborate with smart highways to increase the safety and overall performance under diverse scenarios. We consider the later scenario that needs fine-grained orchestration of all tasks that are executed by the different components in the larger intelligent system~\cite{wang2016towards, rios2016survey}. Intelligent systems rely on artificial intelligence (AI) or machine learning (ML) and need to process the tasks within specified timing constraints. Data intensive AI/ML task processing have traditionally relied on cloud computing~\cite{loven2019edge}. However, the recent emergence of edge computing has introduced a technology that can host the data closer to the devices and allow faster turnaround times for the AI/ML tasks~\cite{li2018learning,zhou2019edge,deng2020edge}. In this paper, we consider task synchronization (i.e., time alignment of task executions) across the different computing nodes for such AI/ML applications. 



The problem of synchronization in real-time AI applications stems from the trade-off between have tightly time-aligned task executions and high throughput (fast execution times). Due to the heterogeneous nature of edge devices, there are new issues that arise from synchronization among the edge workers. Synchronization is important in federated learning where the main challenges are optimization and communication. There is a need for aggregation of local updates. The aggregation process has to be synchronized in order to get good results from the learning process. Current synchronization methods in use include 
Asynchronous Parallel (ASP)~\cite{lian2018asynchronous, keuper2015asynchronous}, Bulk Synchronous Parallel (BSP)~\cite{siddique2016apache}, Stale Synchronous Parallel (SSP)~\cite{ho2013more, cipar2013solving} and Dynamic SSP (DSSP)~\cite{zhao2019dynamic} all of which include aggregating local updates at the server (controller). 

One of the required steps in real-time AI application is data aggregation. Captured data need to be collated, cleaned and pre-processed before training can be done. This process can be made faster using synchronized data capture where the capturing of data across devices is time aligned. Thus, less time is spent in the data pre-processing stage. The quality of data captured is likewise higher. An example is in 
the use of drones in activities such as rescue missions after natural disasters is increasing with recent advancements in drone technologies \cite{valsan2020unmanned}. If there is a need to recreate the search space at a particular point in time, a time skew in the capturing process by the different drones will lead to poor reconstruction of the search area from the snapshots taken by the individual drones.

We aim to develop a fast synchronization scheme with a fast rate of synchronization by minimizing the number of messages required to reach synchronization through the use of a late notification protocol and clustering. The clustering is done such that worker nodes with a high probability of staying tightly synchronized to some bounds are put in the same cluster. To achieve fast synchronization,  we limit the controller involvement in making synchronization decisions.  The unique contributions of this work are as follows:

\begin{enumerate}
    \item We develop a game theoretic model to help in deciding the best synchronization choices and fixing the synchronization options for workers to synchronize.
    \item We develop a disconnection tolerant late notification protocol that maintains system efficiency in the presence of network partitions.
    \item We develop a synchronous scheduling algorithm using clustering that achieves synchronization with the minimal overhead.
    
    \item We implement our algorithm in Ray\footnote{https://docs.ray.io/en/latest/} (a Python framework for building and running distributed applications) and compare it against the ASP, BSP and SSP models.
\end{enumerate}




\section{Background and Related Work}

\subsection{Synchronization in Real-time Artificial Intelligence}
The main synchronization schemes that have been adapted for AI tasks are the ASP, BSP, SSP and DSSP models. Different variations and improvements of these models have been proposed over the years. ASP is the slackest form of synchronization where workers do not have to wait for one another. In BSP, the next iteration only starts after all devices have finished executing the previous iteration. Thus, a barrier is fixed such that there is no progress until all devices get to the barrier. While BSP guarantees total participation in synchronization, it performs poorly in heterogeneous systems where devices can perform the same computations for different amount of time. This causes BSP to suffer a lot from straggling devices.

SSP~\cite{ho2013more} on the other hand attempts to solve the straggler problem by relaxing BSP's strict barrier condition and allowing devices to proceed to the next iteration if the gap between the fastest and slowest device is within a bound - called staleness value. SSP is thus an intermediate solution between the BSP and the asynchronous approach. Although, SSP leads to faster executions and less wait time, the quality of synchronization is reduced compared to BSP by allowing asynchrony.

DSSP~\cite{zhao2019dynamic} was proposed as an improvement on the SSP model for deep neural network training. In DSSP, rather than having static staleness, a value is dynamically selected from a range of values based on real-time processing speeds at runtime. A synchronization controller is used to monitor the progress of the worker nodes. A similar approach to DSSP was proposed in ~\cite{zhang2018adaptive} where a performance monitoring model is used to adjust the synchronization delay threshold.

\subsection{Synchronization in Real-time Systems}

In real time computing systems, synchronization is handled using time-triggered controls where all synchronous activities are executed at some predefined points in time~\cite{rajkumar2012synchronization}. Synchronized clocks are used to achieve synchronization in the systems by making all nodes have a common notion of time. Time synchronization schemes~\cite{yoo2020study, elsts2017temperature} have been developed for IoT to allow devices have a common notion of time. In~\cite{elsts2017temperature}, a time synchronization protocol was proposed to mitigate the effect of temperature change on hardware clocks in IoT networks using time-slotted channel hopping. 

A common and naturally occurring synchronization is the one noticed in fireflies~\cite{ramirez2018fireflies, perez2016firefly, brandner2016firefly}. Male fireflies randomly emit flashes at night and over a period of time, the flashes get synchronized. Analysis of firefly synchronization have been carried out and lots of models have been developed over the years. Pulse coupled oscillators (PCOs) have been used to study the synchronization behaviour in fireflies. PCOs refer to systems with interacting oscillators  that oscillate periodically in time. Synchronization of PCO involves making the individual oscillators emit pulses at the same time.

The synchronization schemes proposed in~\cite{olaniyan2018synchronous} and ~\cite{olaniyan2019multipoint} focused mainly on achieving synchronization without much consideration on the message overhead caused by the constant communication between worker nodes and the controller. The algorithms were based on the task attributes, time and component redundancy as well as localization of worker nodes. The controller was heavily involved in evaluating whether the desired ratio (quorum) of workers are available and in making synchronization decisions. This incurred extra message overhead since all workers need to communicate with the controller before proceeding with synchronization.



\subsection{Game Theoretic Synchronization Approaches}

Synchronization games~\cite{simon2017synchronisation, saukh2018synchronization} have been developed to analyse cases where a set of players need to coordinate their choice within a certain subset of players. A positive payoff is gotten if all the members of a coordinating group choose the same strategy. The utility derived from a synchronization game is dependent on the actions the members of the coordinating group choose.

A game theoretic approach for synchronizing pulse coupled oscillators was proposed in~\cite{yin2011synchronization}. The model developed is an extension of the well-known Kuramoto model for synchronizing systems of oscillators. The game is non-cooperative with oscillators in the system competing against one another. An ``oblivious solution'' was developed where individual oscillators do not have access to the full system state. The oscillators make decisions strictly based on local states and a consistent average value.

\subsection{Comparison of this Work and Related Work}

The SSP and DSSP schemes allow some specified tolerated slack in the synchronization process. However, in our work, we do not allow any slack but allow synchronization to proceed if we have the desired quorum of workers available to synchronize. Unlike DSSP, where workers are expected to have the same or very similar runtimes per iteration, we assume a highly heterogeneous system where runtimes may vary from one iteration to another. The synchronization schemes in ~\cite{olaniyan2018synchronous} and ~\cite{olaniyan2019multipoint} incur extra communication overhead from heavy involvement of the controller. This work achieves fast synchronization by reducing the number of messages sent through clustering and by minimizing the involvement of the controller in making synchronization decisions. 


\section{System Model}

\subsection{Node Model}

\begin{figure}[!t]
  \centering
    \includegraphics[width=0.85\linewidth]{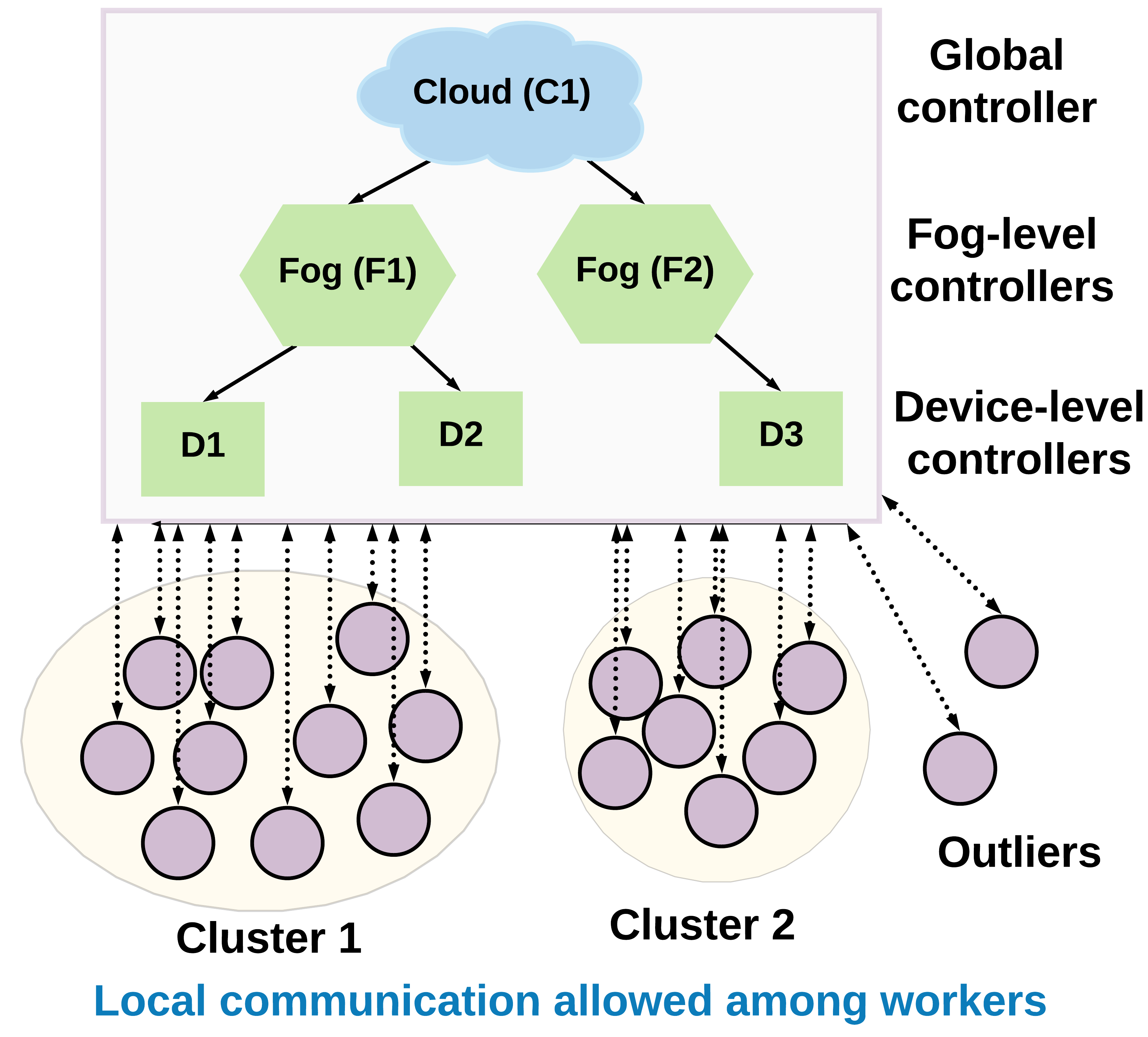}
    \caption{Node model.}
    \label{fig:archi}
\end{figure}

A hierarchical model is used for nodes in our system as shown in Fig.~\ref{fig:archi}. Nodes at the bottom of the hierarchy are called workers. The nodes at upper levels of the hierarchy are called controllers. The controllers could be at three levels - device, fog and cloud levels. Worker nodes can communicate with one another leveraging the underlying fast Wi-Fi local broadcasts. This architecture is suitable for achieving fast synchronization in AI application tasks and edge-based systems because it permits the controller to monitor the progress of workers and allows clustering of workers to reduce message overhead.

Workers are expected to update the controller of their execution progress as they execute a given program. The controller uses these updates to cluster workers. Thus, workers are always part of a logical cluster. However, the clustering details are used only at a synchronization point. The workers know their clusters and the number of workers in each cluster in the system. Tight clock synchronization is assumed across nodes in all levels of the hierarchy. We assume that workers have tightly synchronized clocks.


\subsection{Application Model}

An application written for our system consists primarily of three task types - synchronous, asynchronous and local tasks. All of the three tasks can be at the controller or at the worker. Synchronous and asynchronous tasks are remote calls triggered by the controller on a worker or by a worker on the controller. Local tasks are tasks that are triggered by a node on itself. Synchronous tasks triggered by the controller on workers require that all (or at least a certain ratio of) workers start the execution of the task at the same time. A return result of the execution of a synchronous task to the calling node is expected unlike asynchronous and local tasks. In this work, we focus on remote calls from the controller to workers as it poses the most challenge of coordinating the activities of the workers. Synchronous, asynchronous and local tasks on workers are denoted as $T_{ws}$, $T_{wa}$ and $T_{wl}$ respectively.

We include another task called the progress-tracker task $T_{pt}$. This task is a one-to-one synchronous task on workers where workers report back to the controller on their execution time. Progress tracking is a way for the controller to monitor the progress of worker nodes and to make better clustering and scheduling decisions. Tasks are created such that workers are able to know when they get to half-way point in the execution of the task. This is done such that workers are able to detect if they will be late in finishing the task.

\subsection{Basic Game Model}

We consider a system where nodes can be grouped into clusters such that nodes within a cluster are expected to remain synchronized within some specified bounds. All worker nodes must be connected to a controller to be considered part of the system. We can thus view the controller as a mediator (i.e., trusted third-party). The game is abstracted at the cluster-level, that is, the game is between clusters and the strategies are at the cluster level. However, the utilities derived by workers are strictly based on the worker's participation in the synchronization process. The notations used in this work are shown in Table~\ref{table:notations}.

\begin{table}[!tb]
\caption{Notations.}
\label{table:notations}
\centering
\begin{tabular}{p{1.5cm}p{6cm}}
\hline
\bfseries Symbol & \bfseries Description \\ \hline
$i$  &  Worker $i, i=1,2,\dotsc,N$.\\ 
\hline
$C_k$  &  Each worker is part of a cluster $k, k=1,2,\dotsc,m$. \\  
\hline 
$\left|C_k\right|$  &  Size of cluster $k$. \\  
\hline 
$\left|\mathbb{N}_s\right|$  &  Number of workers that run a sync task. \\  
\hline 
$\omega_{i}(t)$  &  Cost of waiting for $t$ time units by worker $i$ at sync point. \\
\hline 
$t_{av}^{i}$   & Expected available time of worker $i$. \\ 
\hline 
$t_{s}^c$  & Sync time option $c,c=1,2,3$. \\ 
\hline 
$\mathbb{S}_{c}$  &  Utility derived by each worker for running sync task at sync option $c$.\\ 
\hline
$\mathbb{F}_{c}$  &  Cost of aborting sync at sync option $c$.\\ 
\hline
$\mathbb{L}_i(t_l)$  & Utility derived by worker $i$ for executing local task with execution time $t_l$. \\  
\hline 
$\delta(t)$ & Waiting cost before receiving late notification after time $t$. \\
\hline
$\alpha$  &  Worker quorum for synchronization. \\ 
\hline

\end{tabular}
\end{table}


There are three basic choices that can be made by a worker upon getting to a sync point - (i). {\emph wait for sync} and {\emph do not wait for sync} ((ii). due to lateness or (iii). due to late notifications received). The factors to be considered by a node in making a choice include the waiting time, the option of running another task, late notifications and how fast the decision to sync can be made. The controller (mediator) uses the game in making synchronization decisions and scheduling the sync options. The controller specifies sync options based on the choices that maximizes the total utility derived. We assume a silent notification protocol for achieving consensus. Thus, workers proceed to synchronize at the next available synchronization option if they do not get any messages from other workers. The multiple sync options provide workers with the next line of action without the need to send any messages in case a sync option fails. At runtime, late notifications are used to change decisions, it gives a way for workers to skip a particular sync option and choose another option or quit synchronization.

The main components of the game are:

\begin{enumerate}

    \item Player: A strategic decision maker in the context of the game. The two clusters in our case. They make the decisions whether to sync or not at any given sync option.
    
    \item Strategy: The actions of players which include \emph {wait for sync}, \emph {no wait due to lateness} and \emph{no wait due to late notification received}. All players will try to find the best strategy to maximize their payoff.
    
    
\end{enumerate}
\section{Clustering}

We cluster nodes in our system for two main reasons - (i) to reduce the number of messages involved in synchronization, and (ii) to help the controller in making better scheduling decisions. Clustering of workers is done at the controller using the report gotten from workers whenever they get to a reporting point. Thus, the workers are iteratively reporting their execution progress to the controller at each report point. The clustering of workers has to be done such that workers with similar execution progress over time are put in the same cluster. That way, we are expecting that workers within the same cluster will remain tightly synchronized.


We conduct initial experiments to motivate the need for clustering workers and to determine if the choice of having two major clusters is valid. We use the Density-Based Spatial Clustering of Applications with Noise (DBSCAN) clustering algorithm~\cite{kumar2016fast} to group workers into clusters. DBSCAN groups together points that are close to each other based on a distance measurement (usually Euclidean distance) and a minimum number of points. It also marks as outliers the points that are in low-density regions.

We ran four example neural network training tasks using the Python's sklearn \textit{MLPClassifier} and \textit{MLPRegressor} on sklearn's iris dataset continuously on $54$ physical machines. We measure the runtime of each iteration for each task. We collect a total of $6,000$ data points and create clusters using $500$ data points with overlapping data points of $100$ for the next cluster created. We thus have a total of $60$ clustering points. 

\begin{figure*}[!ht]
\begin{multicols}{3}
  \centering
    \includegraphics[width=\linewidth]{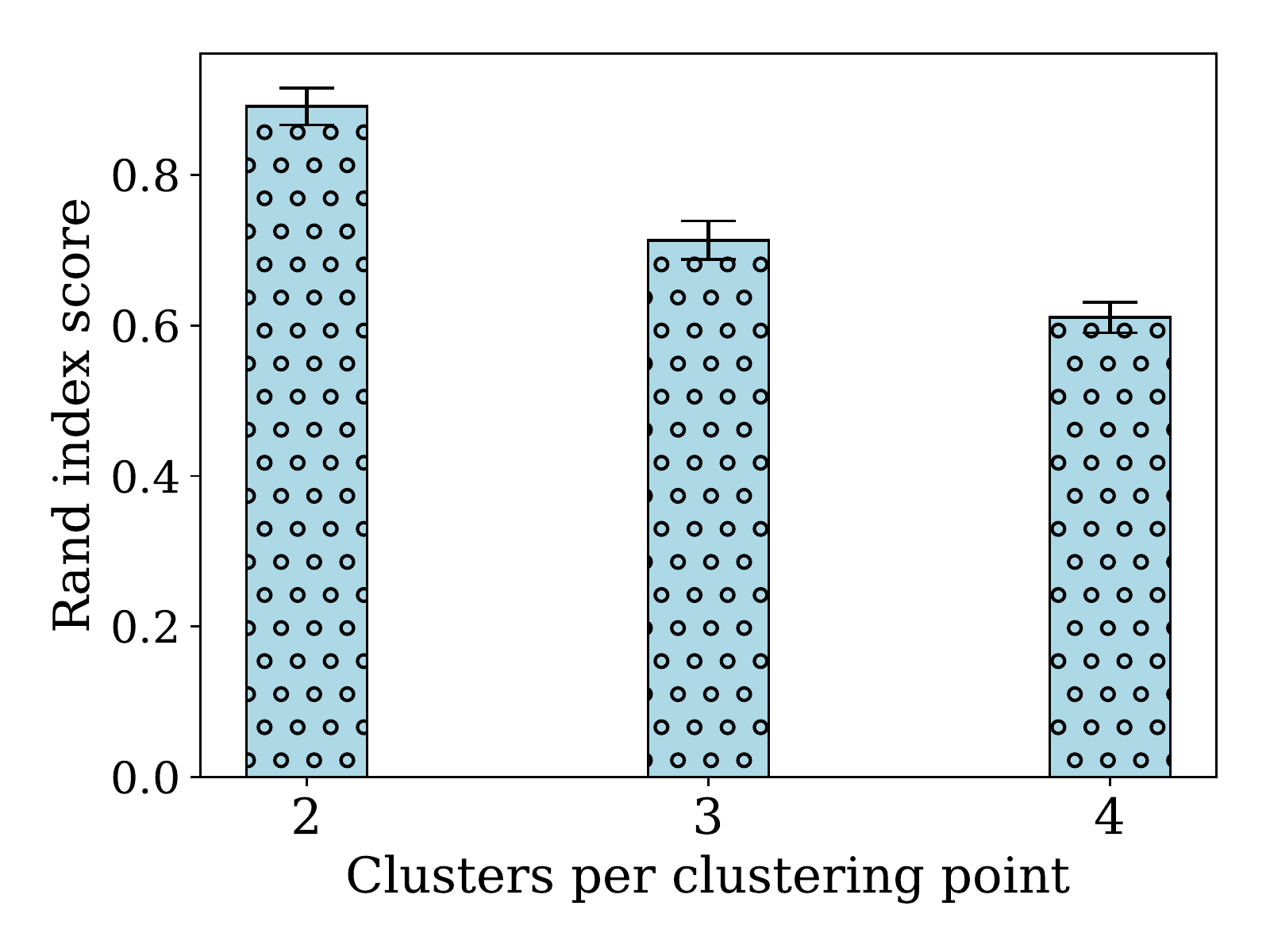}
    \caption{Adjusted Rand Index scores for having $2$, $3$ and $4$ clusters per clustering point.}
    \label{fig:2rand}
    \centering
    \includegraphics[width=0.93\linewidth]{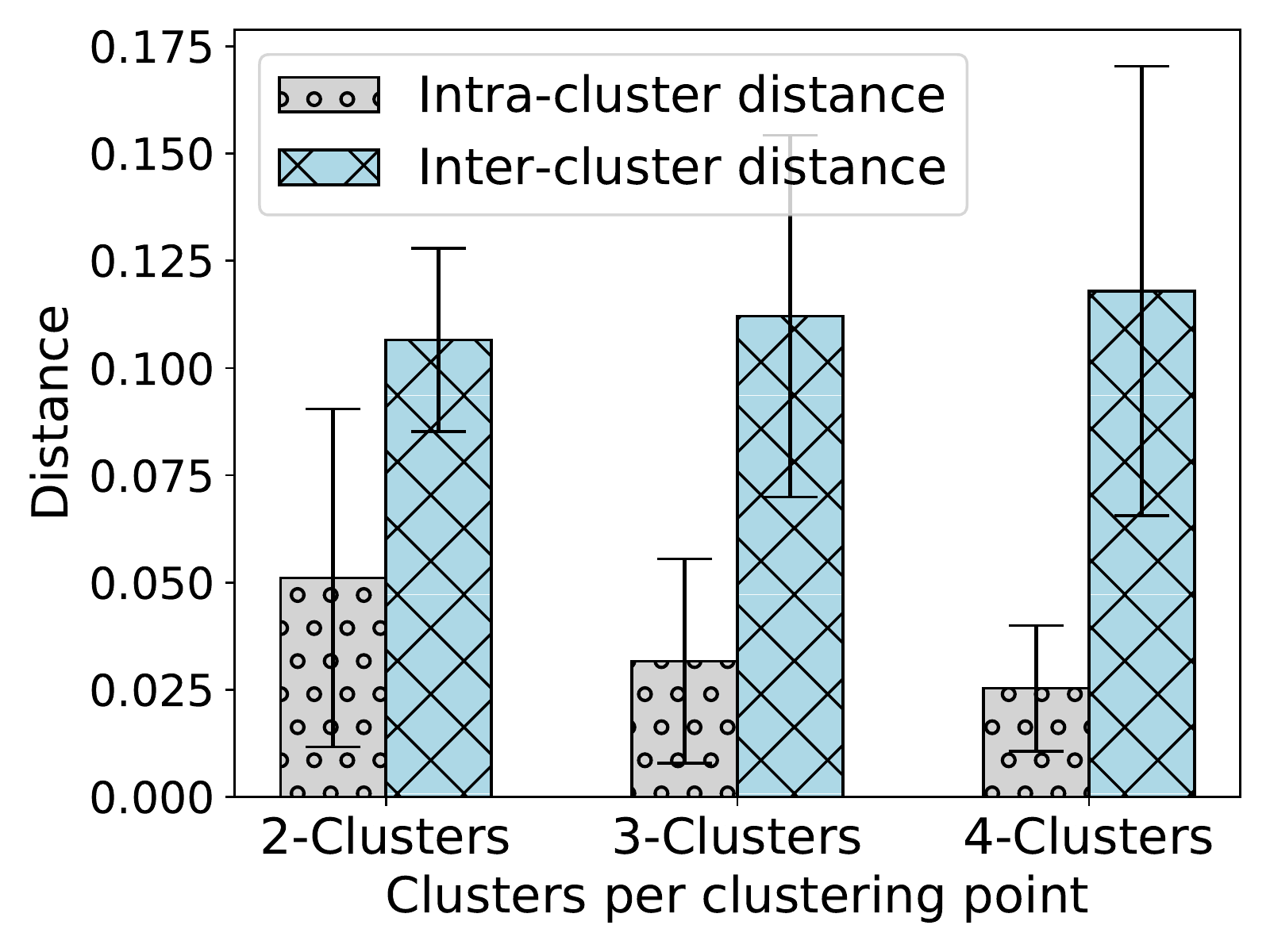}
    \caption{Inter- and intra-cluster distances for $2$, $3$ and $4$ clusters per clustering point.}
    \label{fig:inter_rand}
    \centering
    \includegraphics[width=0.9\linewidth]{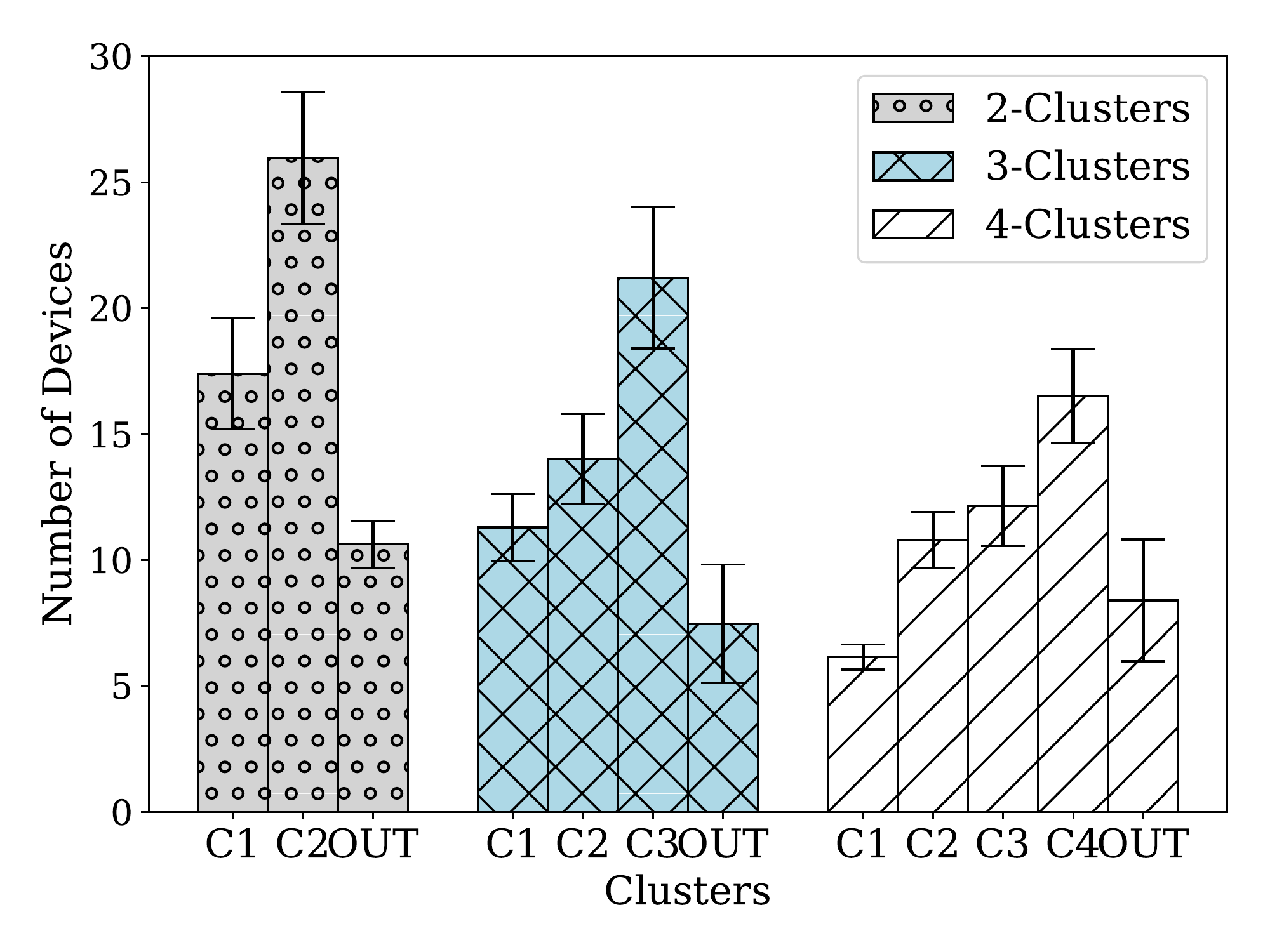}
    \caption{Devices per cluster including outliers for $2$, $3$ and $4$ clusters per clustering point.}
    \label{fig:clus_compare}

\end{multicols}
\end{figure*}

To evaluate the clusters created, we measure the adjusted Rand index score (similarity measure between two clusterings) when $2, 3$ and $4$ clusters are formed  is shown in Fig.~\ref{fig:2rand}. We do an all-to-all comparison of all the clusters formed at each clustering point in evaluating the adjusted Rand index score. This is because more cluster stability is expected when we have two clusters only. As the number of clusters formed increase, there is a higher tendency of machines changing clusters from one clustering point to the other.

The cluster composition and the inter- and intra-cluster distances when $2$, $3$ and $4$ clusters are formed are shown in Fig~\ref{fig:inter_rand} and Fig~\ref{fig:clus_compare} respectively. The average intra-cluster distance for $2$ clusters formed is  higher than those of $3$ and $4$ clusters. The average inter-cluster distance for $2$ clusters is lower than those of $3$ and $4$ clusters per clustering point. This is because for $2$ clusters formed, each of the two clusters is covering a wider range and the maximum distance between machines within a cluster is higher. However, the distances between the two clusters will be reduced compared to $3$ and $4$ clusters per clustering point.

\section{Synchronization as a Game}

\subsection{Game Specification}

We model the game as a non-cooperative extensive-form game that is used to find the optimal strategy of players with regards to synchronization. We chose an extensive-form game  because there are a different number of choices to be made independently but the payoff derived depends on the choice of the other player. To achieve synchronization, the required ratio $\alpha$ of workers(quorum) must be available to run the sync task. The total utility $\mathbb{U}_c$ derived from running a sync task at sync option $c$ is equally divided across all workers regardless of the size of the cluster they are a part of. 
    $$\mathbb{S}_c = \tfrac{\displaystyle \mathbb{U}_c}{\displaystyle \left|\mathbb{N}_s\right|}$$
The total utility derived from running a sync task is a function of how soon the synchronization occurs. An earlier sync option will yield a higher utility compared to other later sync options. Therefore, players have a higher incentive to cooperate better and sooner to maximize their payoff. 
We consider a two-cluster game. The game has a fast and a slow cluster. The fast cluster is the cluster with workers that have shorter execution times and thus generally become available for synchronization earlier than workers in the late cluster. Workers are grouped into clusters based on their execution progress. There could be outliers (workers that do not fall into any of the two major clusters). Outliers can be part of another cluster, but they are not considered in the game. They proceed with the execution schedule and plan created by the outcome of the game.  We keep track of the clusters only at synchronization points. 

\subsection{Execution Time Distributions}

The execution progress of workers is tracked by the controller through checkpoints in the application. The controller creates distributions for the two clusters for the expected finish time of the task before the sync point from their previous run of the application tasks. Each cluster is represented by a mixture distribution of two Gaussian distributions. The first distribution, $\mathbb{D}_{early} = G(\mu_{ea}, \sigma^{2}_{ea})$ represents the early execution times distribution of the cluster while the second distribution, $\mathbb{D}_{late} = G(\mu_{la}, \sigma^{2}_{la})$ represents the late execution times distribution of the cluster. We assume that the distribution of the execution times of local tasks on workers in both clusters are learned. The distribution is mixture of models and defined as $\mathbb{D}^{lo}_{early} = G(\mu_{lo\_ea}, \sigma^{2}_{lo\_ea})$ and $\mathbb{D}^{lo}_{late} = G(\mu_{lo\_la}, \sigma^{2}_{lo\_la})$.


\subsection{Late Notification Protocol}

The main goal of clustering the workers in our system is to reduce the message overhead required in reaching synchronization. Thus, workers in a cluster are expected to remain synchronized and make the same synchronization decisions. The late strategy that involves sending late notifications to inform the other cluster of lateness requires sending messages. To reduce the number of messages required in classifying a cluster as late, we develop a late notification protocol where $3$ messages are expected to be received from workers in a particular cluster for the cluster to be regarded as late. We assume that workers are able to detect when they will be late when they get to $50\%$ of the current task execution based on the predicted finish time of the cluster for that task. 

The first worker in a cluster that detects it will be late sends out a late notification to workers in the other cluster. After the first notification, we set the probability of sending further late notifications to:
$$P_{late} = \tfrac{2}{N-1}$$ 
Thus, if all the workers in a cluster are late, we expect a total of 3 notifications. 

Workers in a cluster could get stuck at a sync option if late notification broadcast messages are lost due to network partitioning. Network partitioning could cause temporary or complete isolation. A worker that is temporarily isolated is only disconnected for one or a few iterations while a completely isolated worker is totally disconnected from other workers in the system. To ensure safety (to prevent a worker or workers from being stuck at a synchronization point) in case of temporary isolation, we embed previous late notifications in new late notifications. Thus, the second late notification will contain the first late notification. If the first late notification gets lost due to network partitioning, workers in the other cluster will get both late notifications embedded in the second notification. A worker that is completely isolated will continue executing tasks in isolation until it gets connected back.

\subsection{Extensive Form of Synchronization Game}

\begin{figure}[!tb]
  \centering
   \includegraphics[width=0.75\linewidth]{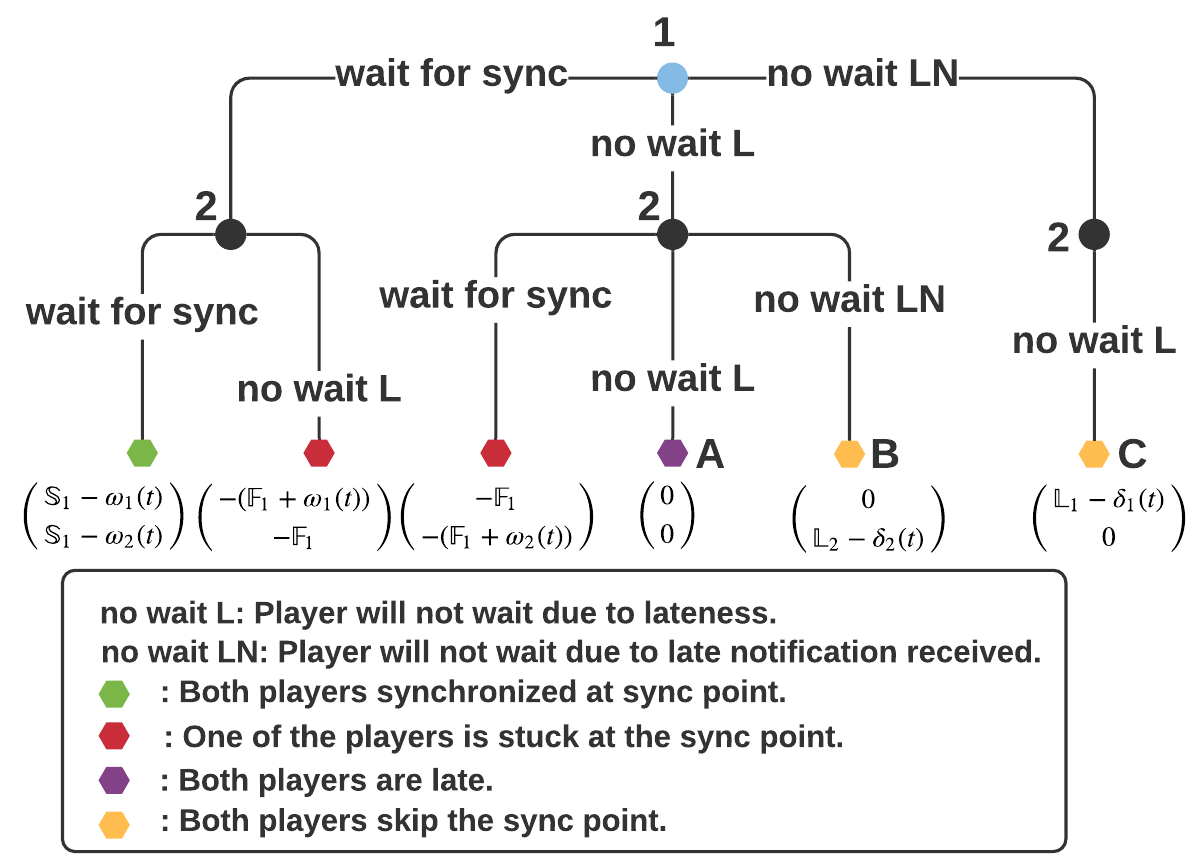}
    \caption{First pass of the extensive form of the two-player synchronization game. Nodes A, B and C are non-terminal nodes where the game proceeds to the second pass.}
    \label{fig:pass1}
\end{figure}

The controller in our system needs to create a static schedule with different sync options and broadcast the schedule to worker nodes. To fix these sync options, the controller uses a game between worker nodes abstracted at the cluster-level. The controller creates the schedule based on the outcome of the game that is expected to yield the maximum payoff. The protocols for choosing which points to synchronize at by the workers is specified by the game. The game is played at the cluster level, although synchronization is done by the workers. If a worker makes a decision, we expect that all the workers within the cluster are going to make the same decision as the cluster. Unexpected behaviour during runtime is tackled using late notifications.
The first pass of the extensive form of the game with two players (clusters) is shown in Fig.~\ref{fig:pass1}. The strategy profile for both clusters include \{\textit{sync, no-sync-late, no-sync-late-notification}\}. In the first pass, there are four possibilities as shown in Fig.~\ref{fig:pass1}; (i) both clusters synchronize at first sync option (green node). (ii) sync aborted because one of the clusters is stuck at first sync option (red node). (iii) sync option skipped because both clusters are late (purple node). (iv) sync option skipped because one cluster got late notification (yellow node).

The second pass is similar to the first pass and it originates from the non-terminal nodes in the first pass. The choices are the same as in the first pass. However, the payoff is cumulative, that is, the payoff derived after the second pass is the addition of that derived in the first pass and second pass. The second pass likewise has $3$ non-terminal nodes as in the first pass. The third and final pass starts with the non-terminal nodes in the second pass. All the exit nodes in the third pass are terminal nodes. The payoff after the third pass is the sum of all the payoffs at all the passes.

We make the following definitions. 

\textbf{Definition 1}: \textit{The utility $\mathbb{S}_1$ derived by a worker from synchronizing at the first option is much greater than the utility $\mathbb{S}_2$ derived from synchronizing at the second option and the third sync option $\mathbb{S}_3$ regardless of any added utility $\mathbb{L}_2$ derived from running a local task.} 

\begin{equation}\label{eqn:syncbound}
\begin{split}
    \mathbb{S}_1 &> \mathbb{S}_2 > \mathbb{S}_3 \\
    \mathbb{S}_1 &> \mathbb{S}_2 + \mathbb{L}_2 > \mathbb{S}_3 + \mathbb{L}_2
\end{split}
\end{equation}

Thus, the earlier synchronization is attempted, the more the payoff that is gotten. The payoff from synchronizing at a later time can never be more than the payoff of synchronizing at an earlier time.

\textbf{Definition 2}: \textit{The cost of aborting sync increases downwards from the first sync option to the third sync option.} 

\begin{equation}\label{eqn:nosyncbound}
    \mathbb{F}_1 < \mathbb{F}_2 < \mathbb{F}_3
\end{equation}

What this means is that it is better to abort synchronization at the first sync option and move on with the execution plan rather than wait till other options to abort sync.





\textbf{Definition 3}: \textit{The strategy (\textit{sync, sync}) at the first sync option is Pareto optimal since there is no other strategy set that gives a higher payoff.}

There are other game strategies that give an optimal solution depending on the runtime operation of the clusters. The strategies (\textit{sync, sync}) at second and third sync options are also optimal solutions to the game depending on what happens at runtime. However, the Pareto optimal solution is the one where both clusters synchronize at the first sync option as evident from Definition 1.

\section{Analysis of the Synchronization Game}

\subsection{Optimal Synchronization Options}

Let ($\mathbb{V}_1, \mathbb{V}_2$) be the payoff vector for cluster 1 (fast cluster) and cluster 2 (slow cluster) respectively and the cumulative payoff $\mathbb{P} = \mathbb{V}_1 + \mathbb{V}_2$. The optimum solution $S^*$ to the game is defined as:

\begin{equation}
    S^* = \arg \min_{\displaystyle t^c_s} \max_{\displaystyle \mathbb{P}} \displaystyle\sum_{i=1}^{2}\mathbb{V}_i
\end{equation}

The highest payoff is gotten when both clusters decide to wait for synchronization at the first sync option. The combination of both strategies by both clusters forms a Nash equilibrium since neither cluster can get a higher payoff by switching to a different strategy as evident in Definition 1. Thus, if one cluster chooses to wait for synchronization, it knows that the other cluster has no incentive to not wait for synchronization. 
To determine the optimal number of sync options, we look at different scenarios in the game. We have two clusters arriving at the sync point; fast and slow cluster. The first choice will be to attempt synchronization at the point where we expect to meet the desired quorum. According to Definition 1 and the payoff gotten from synchronizing ($\mathbb{S}_c - \omega(t))$, a higher payoff is gotten if synchronization is attempted as soon as we have quorum such that $\omega(t_w)$ will be close to $0$. Thus, the first sync option looks to minimize ($\omega(t_{w1}) + \omega(t_{w2})$).

If the slower cluster gets late to the first sync option, there is a need for a second sync option. The second option has to be fixed such that it maximizes the cumulative payoff ($\mathbb{P}$). The earlier cluster will look to get a higher payoff by running a local task. When a cluster is late, it has no incentive not to inform the other cluster by sending a late notification. The optimal option is to fix the second sync option such that if ($\mathbb{L}_1(t_l) > \omega(t_{w2}$), then the fast cluster executes a local task before attempting synchronization again. Else if ($\mathbb{L}_1(t_l) < \omega(t_{w2}$), synchronization is attempted immediately after the late cluster becomes available. This guarantees that the cumulative payoff for both clusters is maximized. 

In a case where the fast cluster executing the local task overshoots the second sync option, it has no incentive not to inform the second cluster. The second cluster can in turn run a local task to improve the cumulative payoff $\mathbb{P}$ if ($\mathbb{L}_2(t_l) > \omega(t_{w2})$) and the third sync option can be fixed after the expected finish time of the local task on the second cluster. Else if running the local task does not improve the cumulative payoff, synchronization can be attempted immediately after. Beyond this point, there is no guarantee that an optimal solution can be found that guarantees a higher cumulative payoff since both clusters would have executed local tasks and there is no other way to improve the cumulative payoff pending synchronization. Thus, it is not an optimal strategy to keep waiting for synchronization beyond this point.

In the case where a cluster is unable to make the first synchronization option, the next optimal solution is to attempt run any local task if available and go to the second synchronization option. The explanation above still stands as no cluster has any incentive to defect from waiting if the other cluster waits. According to Definition 2, a cluster will prefer to wait for synchronization if it expects its local task to overshoot the sync option. In order to get $S^*$, it is imperative to fix the sync options such that the number of expected workers from both clusters is greater than or equal to the desired quorum. Synchronization has to be attempted at the earliest possible options. 

\subsection{Fixing the Synchronization Options}

Given that we have the mixture distributions $\mathbb{D}^1$ and $\mathbb{D}^2$ for the execution times of the fast and slow clusters $C_1$ and $C_2$ respectively, and likewise the distribution of the expected execution time of local tasks on both clusters, we can fix the three synchronization options. The expected available times of the clusters can be chosen from the mixture distributions by choosing the desired percentile $p(x)$. The percentile values are used because we expect workers in a cluster to make similar decisions. 

The sum Z of two normally distributed independent random variables $X (= G(\mu_{x}, \sigma^{2}_{x}))$ and $Y (= G(\mu_{y}, \sigma^{2}_{y}))$ is also normally distributed, $Z = G(\mu_{x} + \mu_{y}, \sigma^{2}_{x} + \sigma^{2}_{y})$.\\

\textbf{First Synchronization Option}:
For the first sync option, we are interested in getting the time value $t^1_s$ such that the desired percentile $p$ on both clusters is available and the desired quorum is met. The percentile is sampled from the early execution distribution $\mathbb{D}^{1}_{early}$ and $\mathbb{D}^{2}_{early}$ for both clusters. We use the early execution distributions in order to fix the first synchronization option as early as possible. Let $X_1$ and $X_2$ be the time values that correspond to the chosen percentiles on both distributions for both clusters. The time $t^1_s$ for the first synchronization option can be fixed by solving the following equation:
\begin{equation}
    \begin{aligned}
        & \text{minimize} 
        && \displaystyle t^1_s \\
        & \text{subject to} 
        && p(x) \left| C_1\right| + p(x) \left| C_2\right| \geq \alpha N, \\
        &&& X_1 = p(x)\{\mathbb{D}^{1}_{early}\},\\
        &&& X_2 = p(x)\{\mathbb{D}^{2}_{early}\},\\
        &&& t^1_s = max(X_1, X_2) \\
    \end{aligned}
    \label{eqn:option1}
\end{equation}

\textbf{Second Synchronization Option}:
The second sync option is fixed such that the faster cluster, say $C_1$ either waits for the slower cluster, say $C_2$ (which is late) or executes a local task (if available) if it increases the cumulative payoff. The percentile for the expected available time $t^{2}_{av}$ is drawn from the distribution $\mathbb{D}^{2}_{late}$. The second sync option $t^2_s$ is gotten by solving the equation:
\begin{equation}
    \begin{aligned}
        & \text{minimize}\quad \displaystyle t^2_s \\
        & \text{subject to} \\
        & p(x_1) \left| C_1\right| + p(x_2)\left| C_2\right| \geq \alpha N, \\
        &  X^{'}_1 = X_1 & \text{if} \quad  \mathbb{L}_1(t^1_l) < \omega(t_{w2}),\\
        &  X^{'}_1 = X_1 + p(x_1)\{\mathbb{D}^{1}_{early}  + \mathbb{D}^{lo1}_{early}\} & \text{if} \quad  \mathbb{L}_1(t^1_l) \geq \omega(t_{w2}),\\
        &  X^{'}_2 = p(x_2)\{\mathbb{D}^{2}_{late}\},\\
        &  t^2_s = max(X^{'}_1, X^{'}_2)
    \end{aligned}
    \label{eqn:option2}
\end{equation}

$X^{'}_1$ is the time point where we expect a certain percentile of the workers in the faster cluster to be available to synchronize. If cluster $C_1$ executes a local task, $X^{'}_1$ is gotten by getting the desired percentile from the sum of the distributions $\{\mathbb{D}^{1}_{early}$ and $\mathbb{D}^{lo1}_{early}\}$.

\textbf{Third Synchronization Option}:
The last synchronization option is fixed to cater for the situation where the cluster ($C_1$) running the local task is late to the second sync option and sends a late notification to cluster $C_2$. The other cluster $C_2$ can decide to wait or run a local task. This is dependent on which of the choices increases the cumulative payoff. The new expected available time of $C_1$ is drawn from the distribution $\mathbb{D}^{lo}_{late}$. $t^3_s$ is fixed by solving:

\begin{equation}
    \begin{aligned}
        & \text{minimize} \quad \displaystyle t^3_s \\
        & \text{subject to} \\
        & p(x^{'}_1) \left| C_1\right| + p(x^{'}_2)\left| C_2\right| \geq \alpha N, \\
        & X^{''}_1 = X_1 + p(x^{'}_1)\{\mathbb{D}^{1}_{early} + \mathbb{D}^{lo1}_{late}\},\\
        & X^{''}_2 = X^{'}_2 & \text{if} \quad  \mathbb{L}_2(t^2_l) < \omega(t_{w1}),\\
        & X^{''}_2 = X^{'}_2 + p(x^{'}_2)\{\mathbb{D}^{2}_{late} + \mathbb{D}^{lo2}_{early}\} & \text{if} \quad  \mathbb{L}_2(t^2_l) \geq \omega(t_{w1}),\\
        & t^3_s = max(X^{''}_1, X^{''}_2) \\
    \end{aligned}
    \label{eqn:option3}
\end{equation}

$X^{''}_1$ is the time point where we expect a certain percentile of the workers in the faster cluster to have finished executing the local task. We switch to the late local task execution distribution $\mathbb{D}^{lo1}_{late}$ since the cluster is late. $X^{''}_1$ is drawn from the sum of the distributions $\{\mathbb{D}^{1}_{early}$ and $\mathbb{D}^{lo1}_{late}\}$. If cluster $C_2$ executes a local task, $X^{''}_2$ is drawn from the sum of the distributions $\{\mathbb{D}^{1}_{early}$ and $\mathbb{D}^{lo1}_{early}\}$.

\subsection{Putting it All Together}

The synchronization algorithm is shown in Algorithm~\ref{algo:sync_option}. The input to the algorithm is the set of two clusters $C = \{C_1, C_2\}$ representing the fast and slow clusters. The controller fixes the $3$ sync options by solving Equations~\ref{eqn:option1} -~\ref{eqn:option3} in that order. 


\begin{algorithm}[htbp]
\SetAlgoLined


\textbf{Controller}:

Fix the three sync options $t^1_s$, $t^2_s$ and $t^3_s$ by solving  Equations~\ref{eqn:option1}, ~\ref{eqn:option2} and ~\ref{eqn:option3} respectively

\textbf{forall} workers \textbf{do}:

\quad \textbf{First sync option}:

\tab \textbf{if} ($t^1_{av} \leq X_1$) \textit{and} ($t^2_{av} \leq X_2$):

\stab $execute(T_{sync}, t^{1}_s$);

\tab \textbf{elif} ($t^1_{av} \leq X_1$) \textit{and} ($t^2_{av} > X_2$) \textit{and} \textit{send($C_2$, late\_notify)}:

\stab \textbf{if} $t^1_l \leq t^{1}_s - t^1_{av}$:

\mtab $execute(T_{local}, t^{1}_{av}$);

\stab proceed to \textit{line 18};

\tab \textbf{elif} ($t^1_{av} > X_1$) \textit{and} ($t^2_{av} > X_2$):

\stab proceed to \textit{line 18};

\tab \textbf{elif} ($t^1_{av} \leq X_1$) \textit{and} ($t^2_{av} > X_2$) and \textit{no\_late\_notify}:

\stab \textit{abort(sync)};

\quad \textbf{Second sync option}:

\tab \textbf{if} ($t^{1}_{av'} \leq X^{'}_1$) \textit{and} ($t^{2}_{av'} \leq X^{'}_2$):

\stab $execute(T_{sync}, t^{2}_s$);

\tab \textbf{elif} ($t^{1}_{av'} > X^{'}_1$) \textit{and} ($t^{2}_{av'} \leq X^{'}_2$) \textit{and} \textit{ send($C_1$,late\_notify)}:

\stab \textbf{if} $t^2_l \leq t^{2}_s - t^2_{av}$:

\mtab $execute(T_{local}, t^{2}_{av}$);

\stab proceed to \textit{line};

\tab \textbf{elif} ($t^1_{av} \leq X_1$) \textit{and} ($t^2_{av} > X_2$) and \textit{no\_late\_notify}:

\stab \textit{abort(sync)};

\quad \textbf{Third sync option}:

\tab \textbf{if} ($t^{1}_{av''} \leq X^{''}_1$) \textit{and} ($t^{2}_{av''} \leq X^{''}_2$):

\stab $execute(T_{sync}, t^{3}_s$);

\tab \textbf{elif} ($t^{1}_{av''} \leq X^{''}_1$) \textit{and} ($t^{2}_{av''} > X^{''}_2$):

\stab \textit{abort(sync)};

\caption{Synchronization algorithm}
\label{algo:sync_option}
\end{algorithm}

\section{Experiments, Results and Discussions}

\subsection{Simulation Configuration}

We use a task graph (DAG) with a mixture of synchronous, asynchronous and local tasks. The task graph is similar to those used in Bulk Synchronous Parallel (BSP), Stale Synchronous Parallel (SSP)~\cite{ho2013more} and Dynamic Stale Synchronous Parallel (DSSP)~\cite{zhao2019dynamic} approaches for synchronizing parameter updates in distributed machine learning and neural networks. The models usually have the following four steps. (i) Compute gradients using local weights. (ii) Push gradients to parameter server to compute global weights. (iii) Pull new computed global weights from parameter server. (iv) Update local weights using global weights.


These models assume that workers are only involved in the model training and updating process. However, in our work, we consider the case where workers are not only involved in model training, but also in the data capture process and usage of the model's output. We introduce local tasks to show activities where the workers need to do some personal computations for effective functioning of the running application. Local tasks are triggered at runtime based on the application's needs and configurations. 

The execution times of a single task on workers is based on a mixture distribution. One for the fast execution and the other for slow execution. The execution times used in the simulations are gotten from traces from the clustering experiments. The times are split into two to depict short ($\mu = 25ms$) and long tasks ($\mu = 80ms$).

The parameters in the simulations are as follows. (i) \textit{Synchronization degree}: ratio of the total machines required to pass quorum. (ii) \textit{Worker size}: The maximum number of workers present in the system at any point in time. (iii) \textit{Simulation rounds}: The number of times the task graph is continuously run. (iv) \textit{Clustering frequency}: This is the rate at which re-clustering is done by the controller.

\subsection{Default Parameter Values and Measurements}

The following parameters are fixed in the simulations unless otherwise stated. The number of independent runs of each simulation is $100$ while each task graph is continuously run in each simulation for $200$ times (rounds). Worker-worker message cost is set at ($\mu = 2ms, \sigma = 0.3$)  and worker-controller message cost is set at ($\mu = 25ms, \sigma = 2$). The synchronization degree is fixed at $0.7$. Local tasks execution times vary from $5ms$ to $10ms$. Clustering cost is set at $20ms$. The same task graph is run on all workers. 

\begin{figure*}[!ht]
\begin{multicols}{3}
  \centering
      \includegraphics[width=\linewidth]{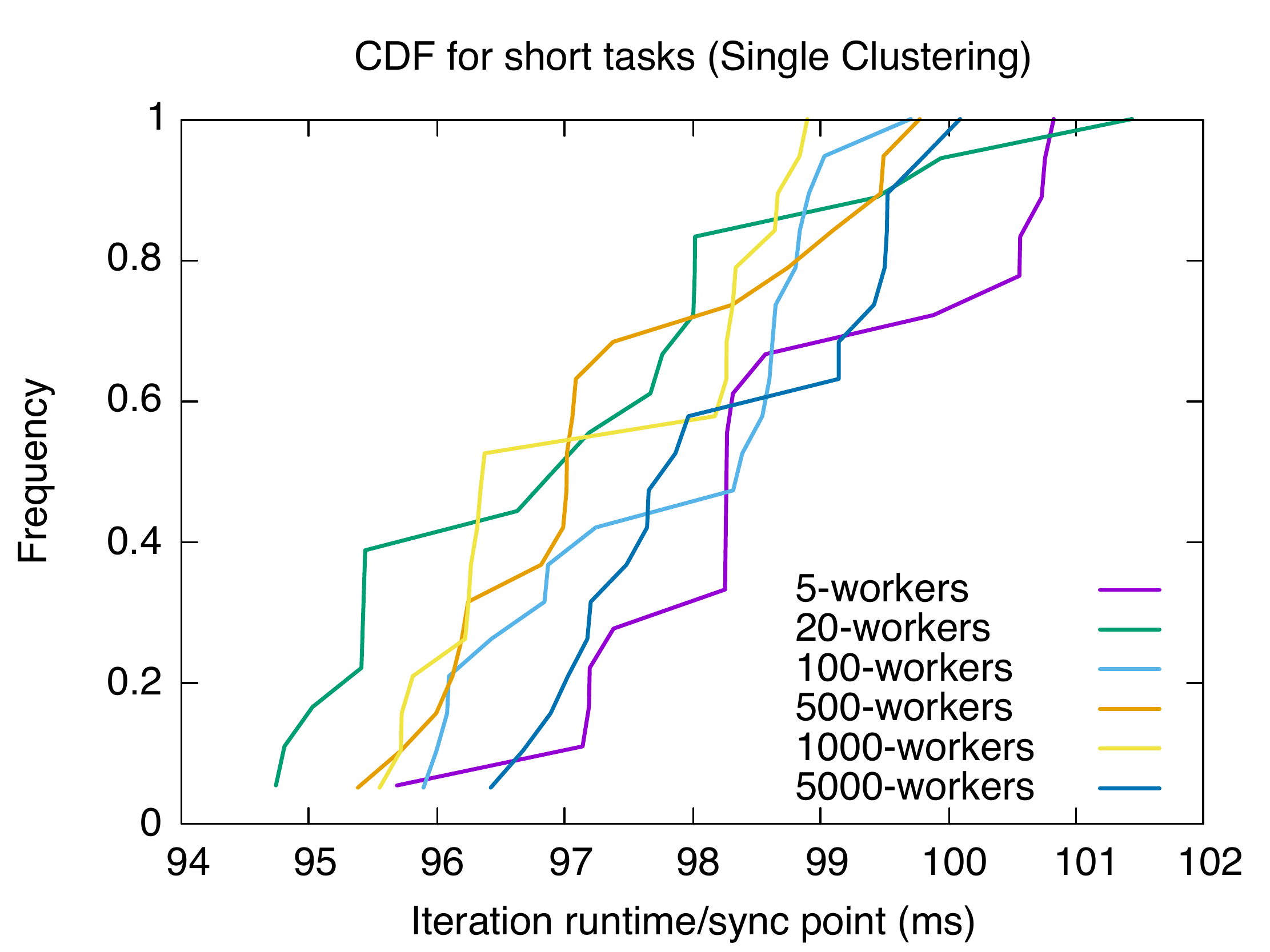}\par
    \caption{Runtime for short tasks (fixed clustering).}
    \label{fig:fixed_runtime}
    \includegraphics[width=\linewidth]{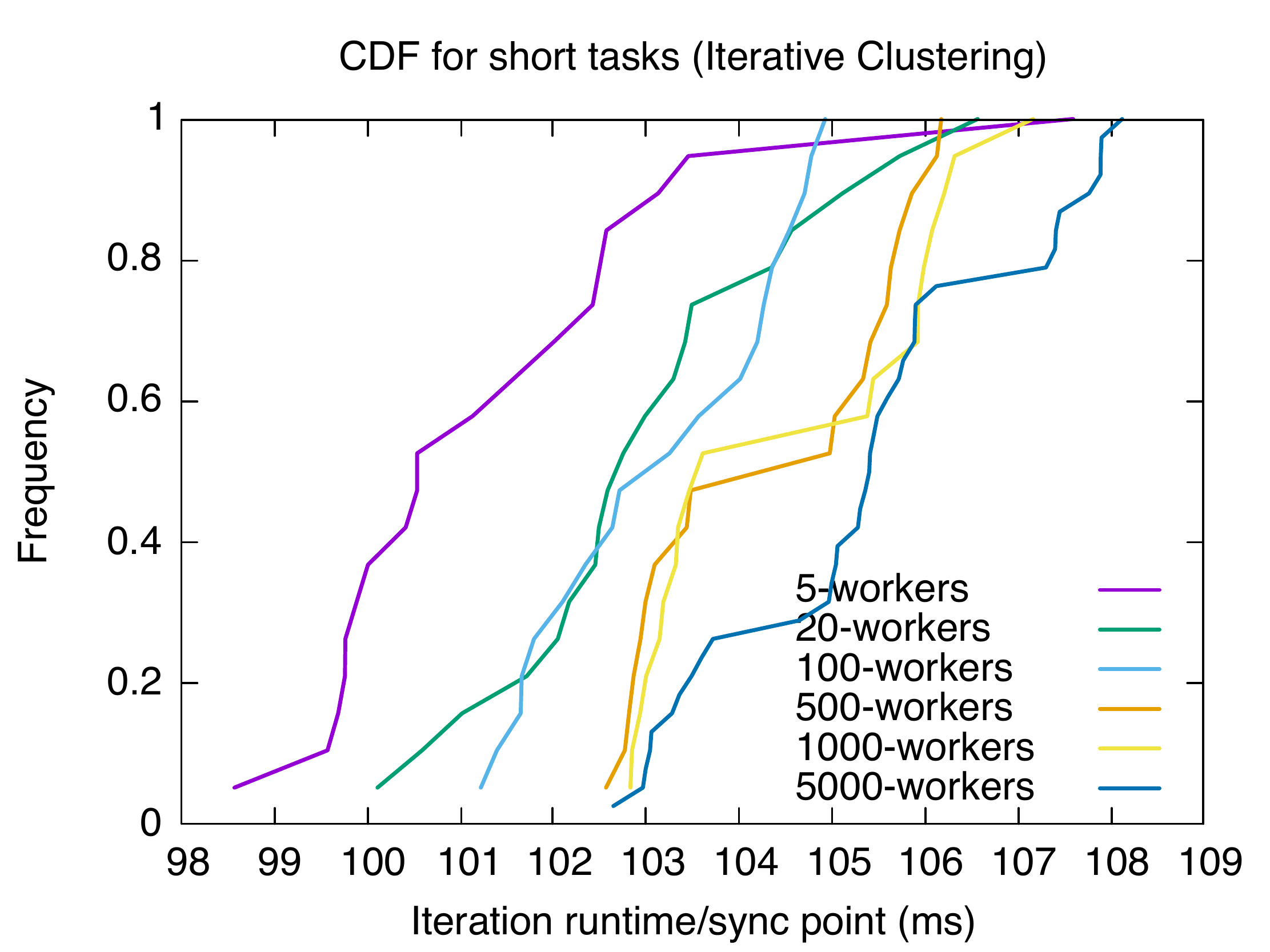}\par
    \caption{Runtime for short tasks (flexible iterative clustering).}
    \label{fig:flex_runtime}
    \includegraphics[width=\linewidth]{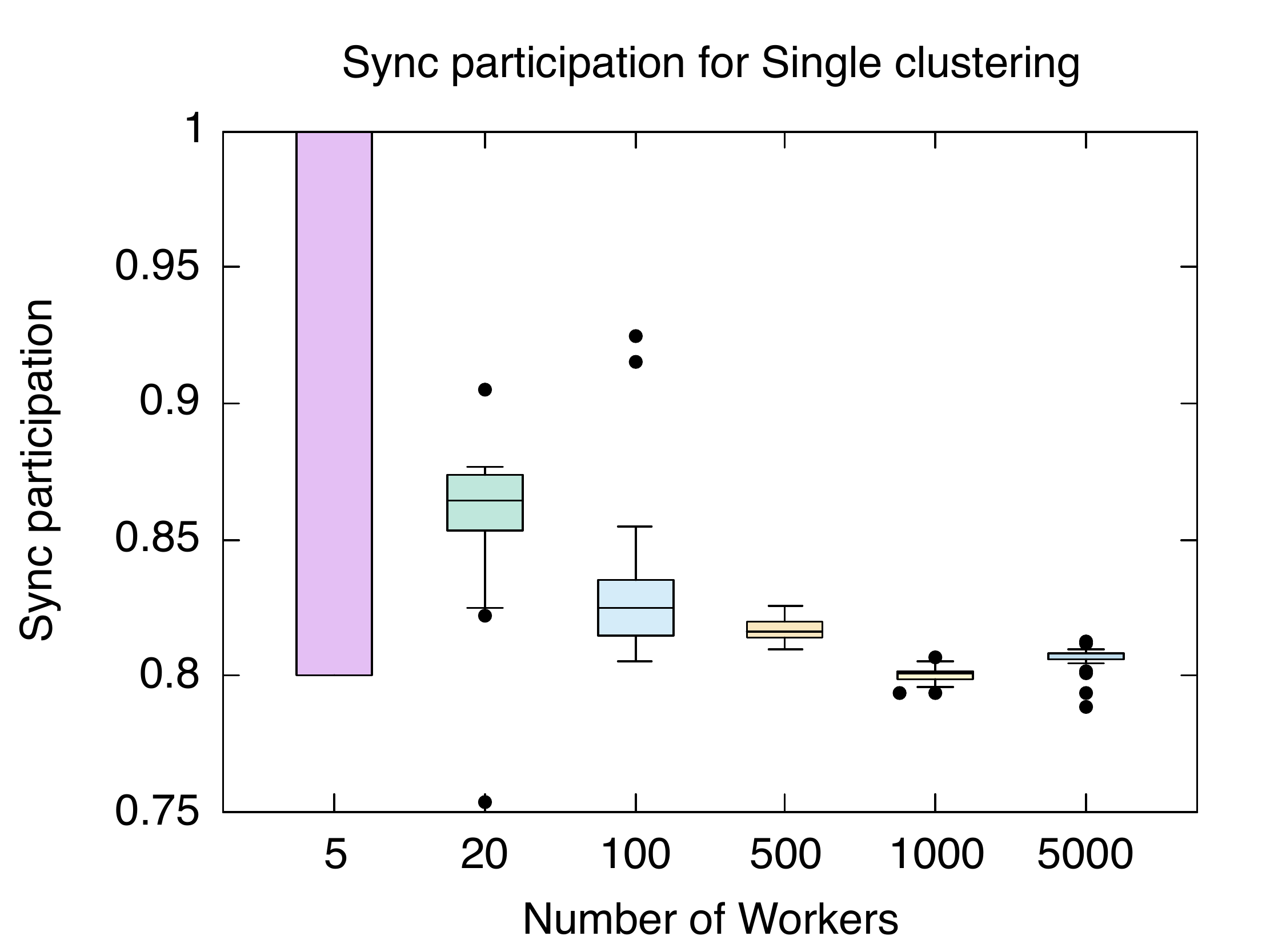}\par
    \caption{Ratio of workers that synchronized for short tasks (fixed clustering).}
    \label{fig:fixed_quorum}
\end{multicols}
\end{figure*}

\begin{figure*}[!ht]
\begin{multicols}{3}
  \centering
    \includegraphics[width=\linewidth]{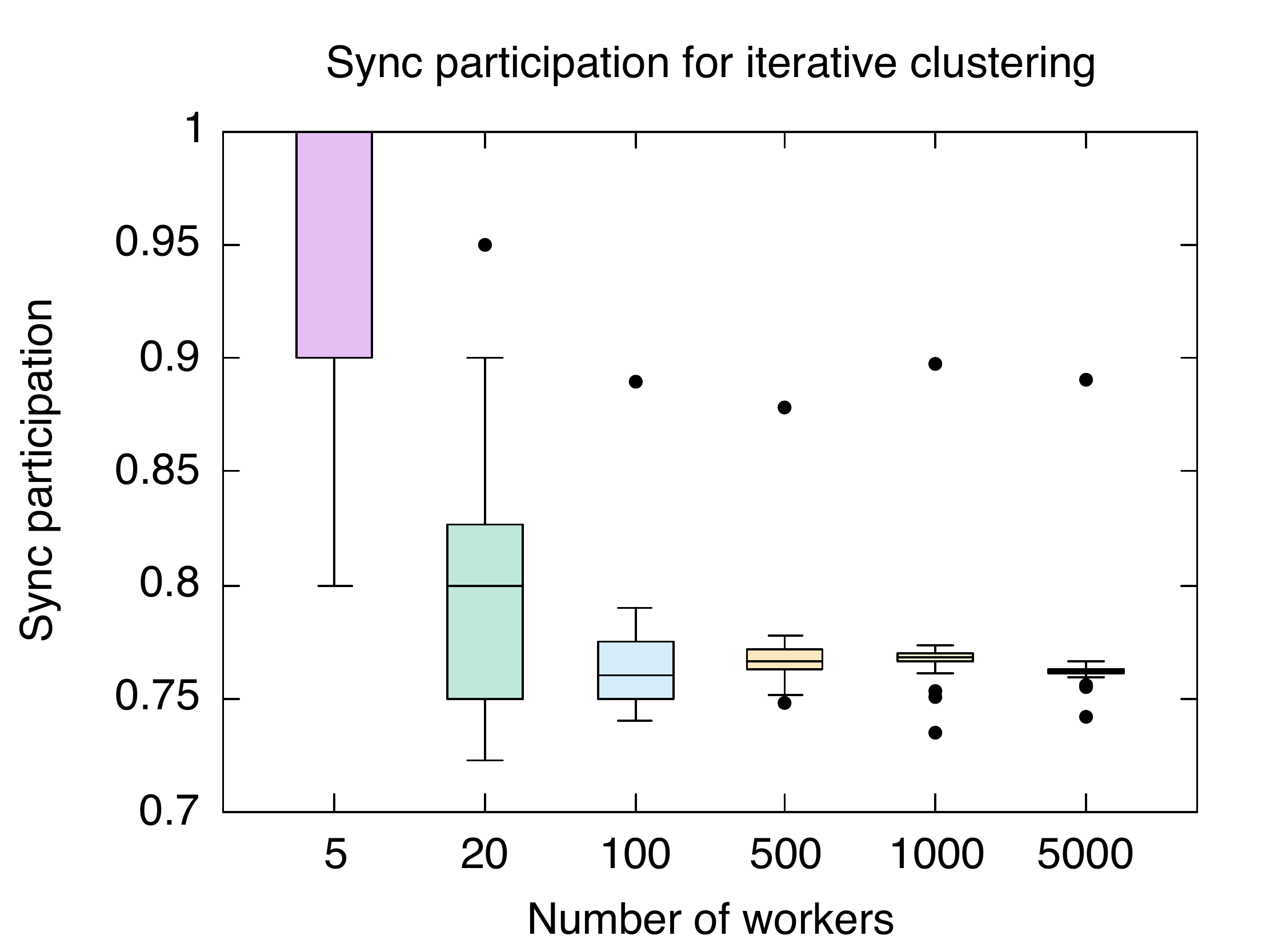}\par
    \caption{Ratio of synchronized workers for short tasks (flexible iterative clustering).}
    \label{fig:flex_quorum}
    \includegraphics[width=\linewidth]{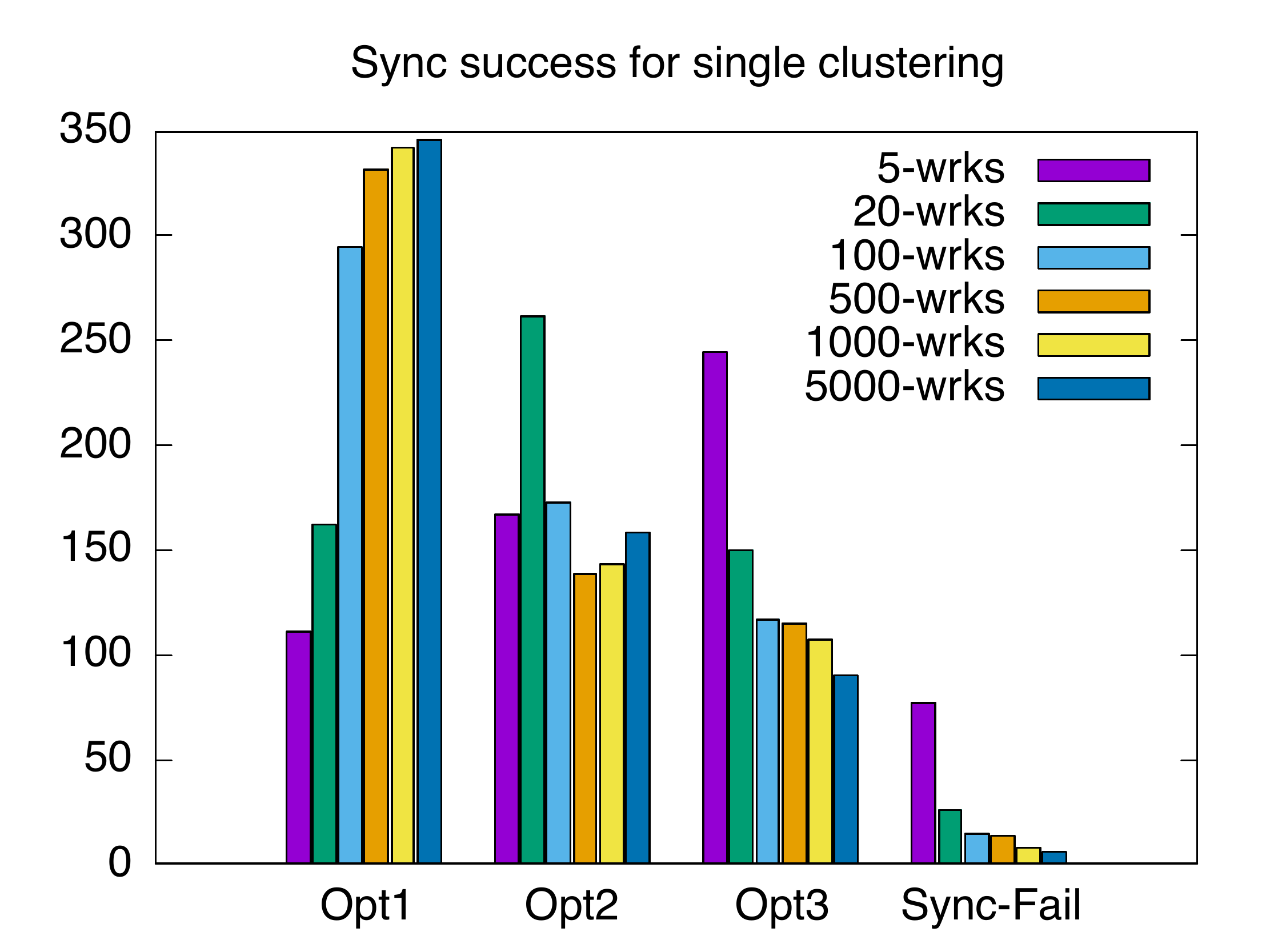}\par
    \caption{Number of successful and failed synchronizations at different sync options (fixed clustering).}
    \label{fig:fixed_succ}
    \includegraphics[width=\linewidth]{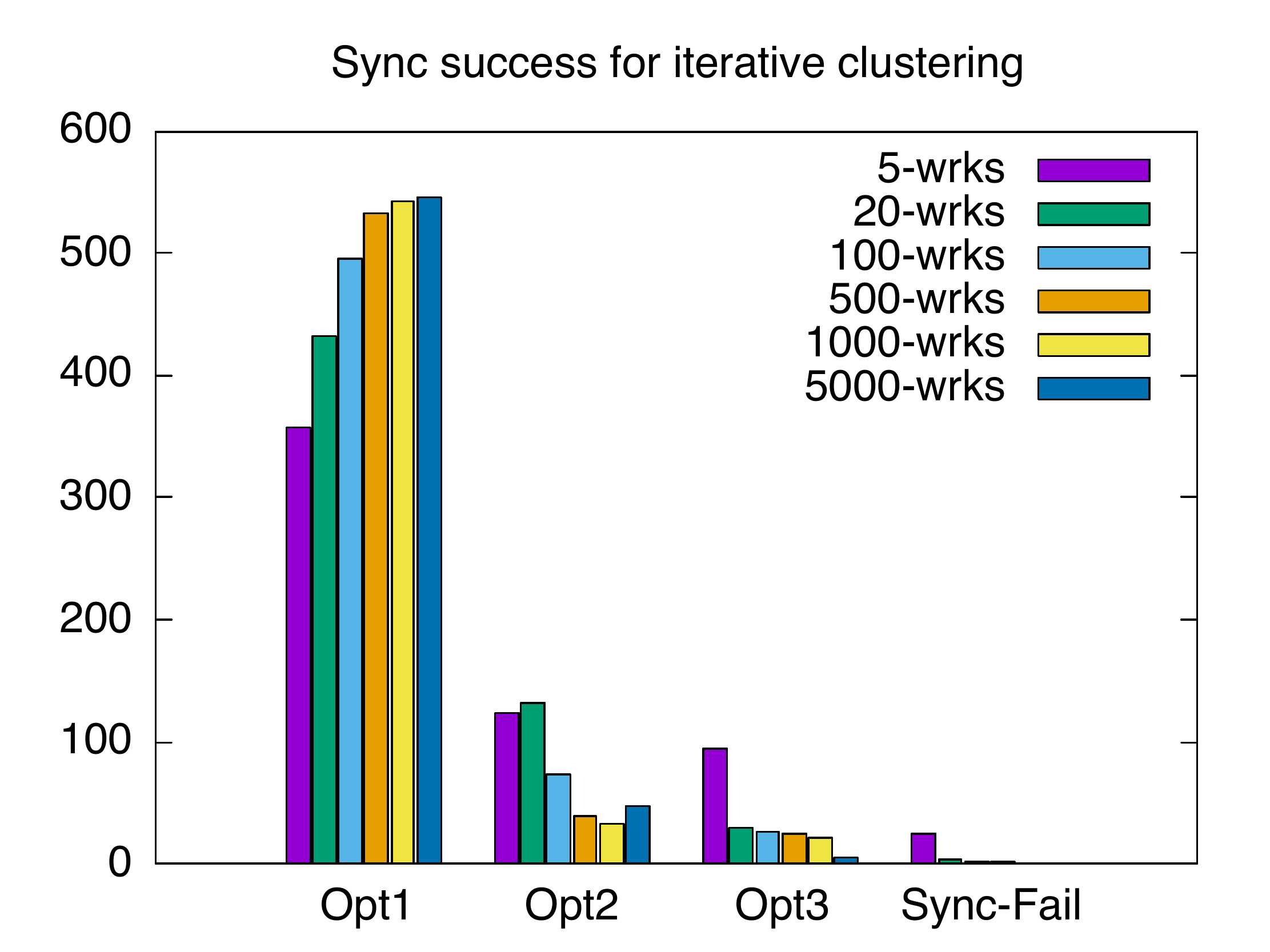}\par
    \caption{Number of successful and failed synchronizations at different sync options (flexible clustering).}
    \label{fig:flex_succ}
\end{multicols}
\end{figure*}

\begin{figure*}[!t]
\begin{multicols}{3}
  \centering
    \includegraphics[width=\linewidth]{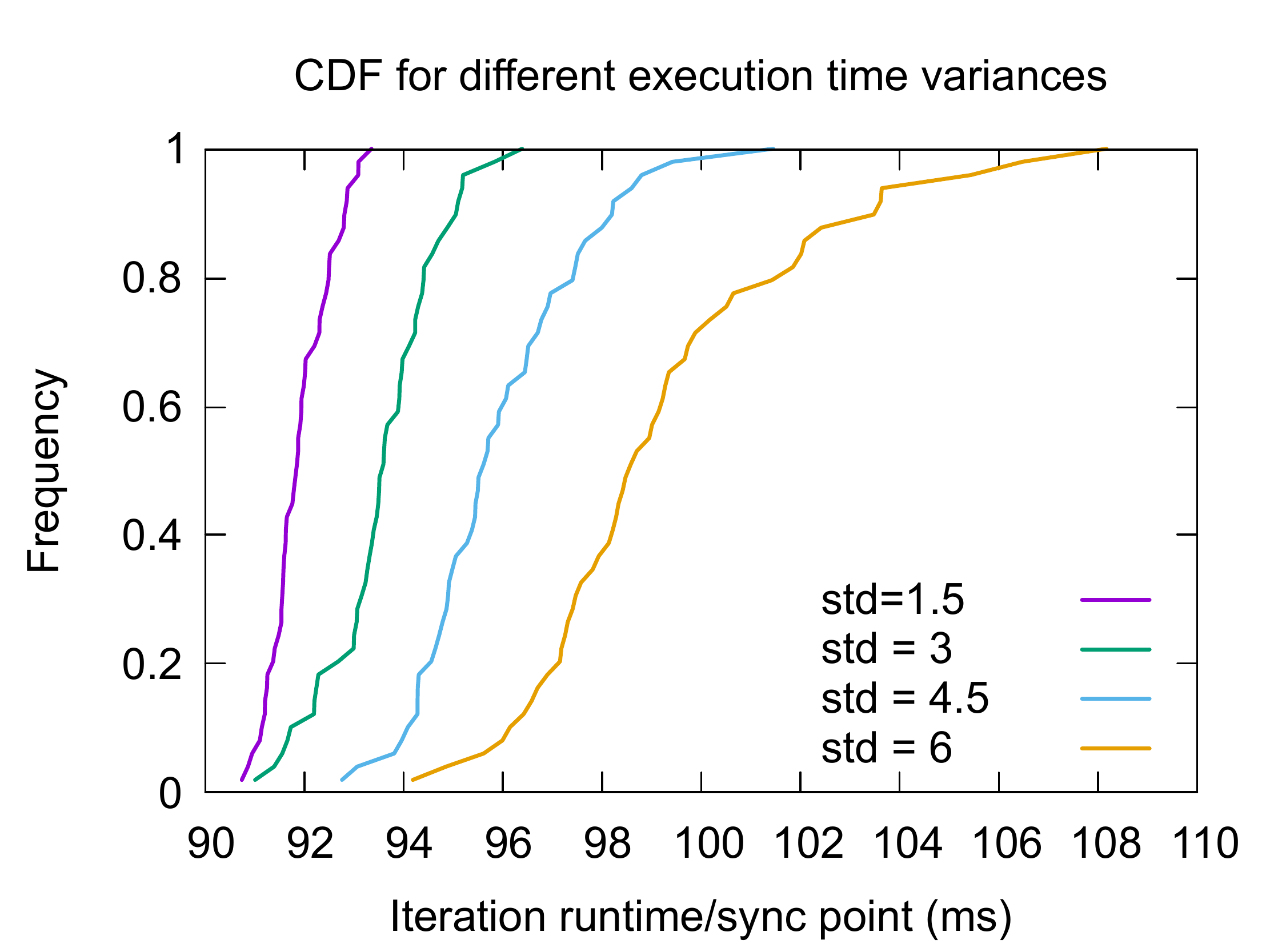}\par
    \caption{Runtime for different task execution time variances.}
    \label{fig:runtime_var}
    \includegraphics[width=\linewidth]{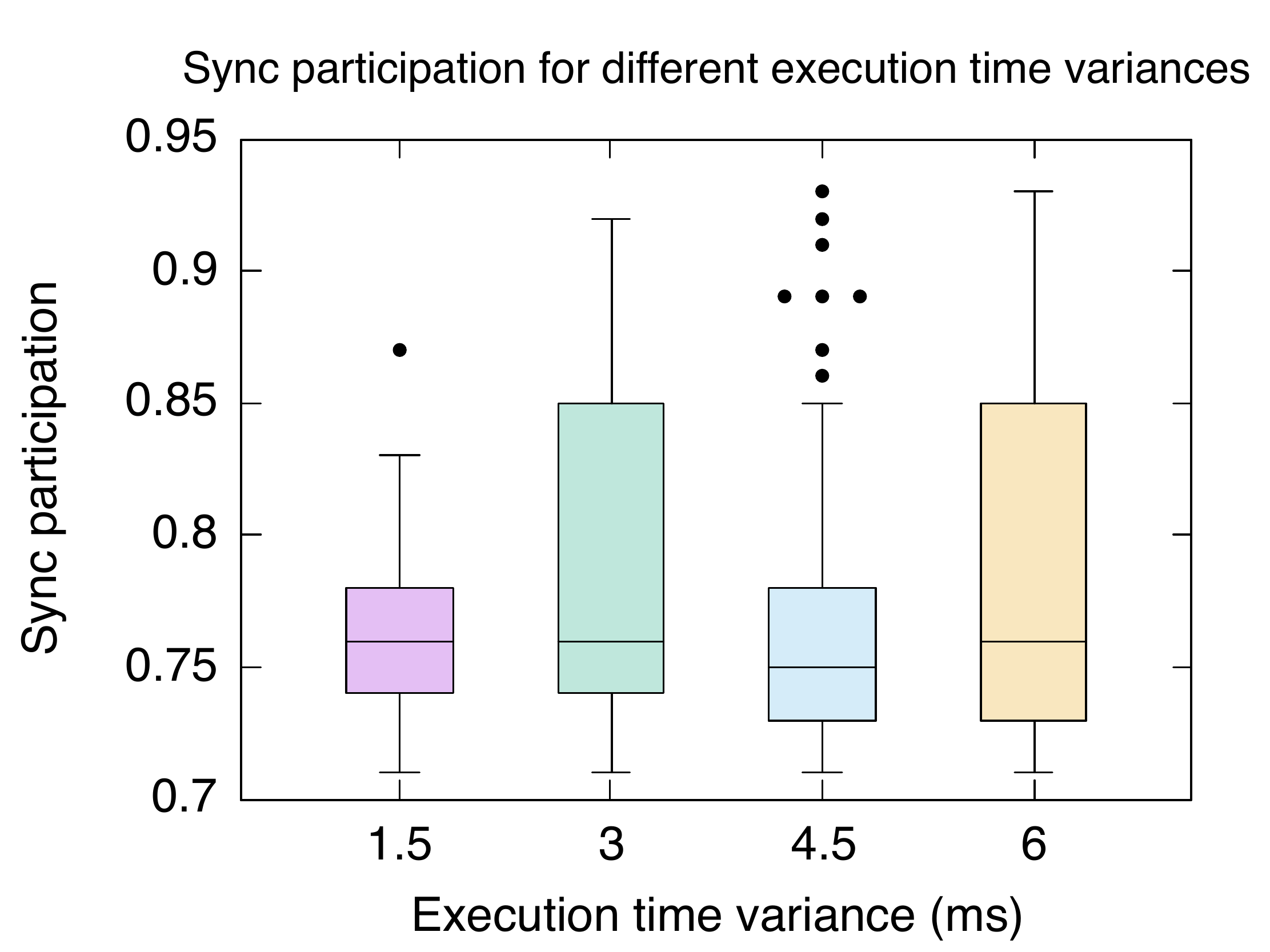}\par
    \caption{Ratio of synchronized workers for different task execution time variances.}
    \label{fig:quorum_var}
    \includegraphics[width=\linewidth]{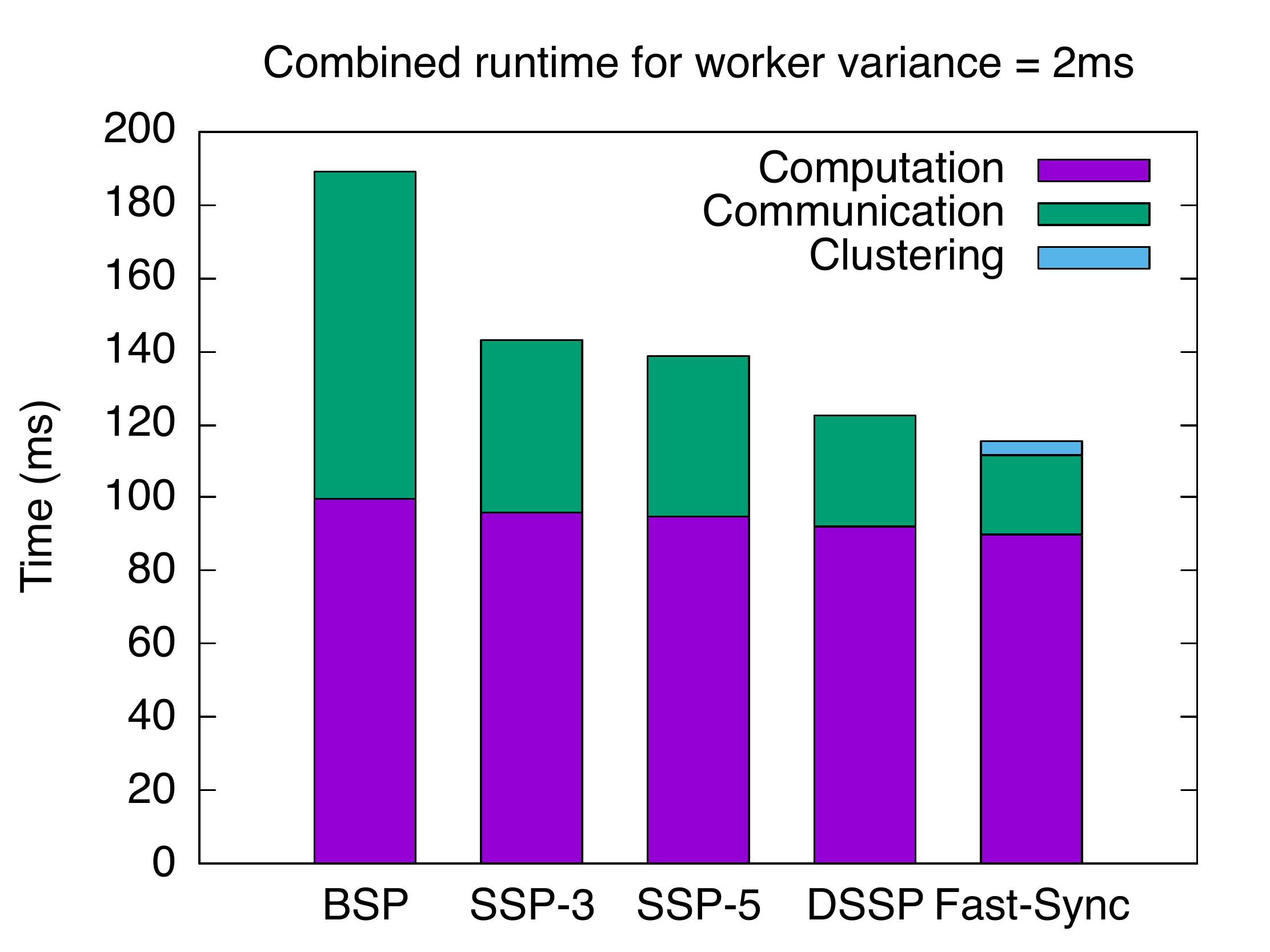}\par
    \caption{Average computation and communication times for worker execution variance of $2ms$ with $20$ workers.}
    \label{fig:combined_plot_var_2}
\end{multicols}
\end{figure*}

  



The following parameters are measured in the simulations. (i) \textit{Runtime/sync point}: the time taken for a single iteration of a task graph divided by the number of sync points. (ii) \textit{Sync success/failure}: the total number of times synchronization was successful or failed at different synchronization options.  (iii) \textit{Sync participation}: the ratio of the total devices that synchronized at a sync point.

\subsection{Simulation Results and Discussions}

\subsubsection{Single vs Flexible Clustering}

We measure the impact of re-clustering on the runtime per sync point, quorum participation, sync success at different options and sync failure. We consider single (fixed) clustering where workers are clustered only once in the system. Thus, workers belong to the same cluster all through the execution. We likewise consider the case where clustering is done after a number of iterations (set to $5$).

Fig.~\ref{fig:fixed_runtime} and~\ref{fig:flex_runtime} shows the runtime per sync point for single and iterative clustering for varying number of workers respectively. The runtime per sync point for single clustering is smaller compared to that for iterative clustering. This is due to extra cost incurred in re-clustering. However, iterative clustering has more sync successes at the first sync option compared to single clustering as well as less failed synchronizations as shown in Fig.~\ref{fig:fixed_succ} and~\ref{fig:flex_succ}. This is because the schedule generated by the cluster using the execution progress distributions of the clusters is updated as re-clustering is done. Thus, improving the accuracy of the schedule. Single clustering has more sync participation than iterative clustering for varying number of workers as seen in Fig.~\ref{fig:fixed_quorum} and~\ref{fig:flex_quorum}. This is because more workers are expected to be available at the second and third sync options as there are more sync successes at those options for single clustering.

\subsubsection{Worker Heterogeneity}

To measure the effect of heterogeneity of workers on our algorithm, we vary the execution time deviation of tasks across workers and explore the impact it has on runtime per sync point and quorum participation. Increasing the standard deviation of a task among several workers increases the possibilities of having stragglers. The task execution time deviation is varied from $1.5ms$ to $6ms$ for $100$ workers and short tasks. Fig.~\ref{fig:runtime_var} shows that the runtime per sync point and the deviation increases as the variance of task execution times across workers is increased from $1.5ms$ to $6ms$. The average sync participation for all execution time variances is about $0.75$ with execution time variance of $1.5$ having a slightly higher average. As the execution time variance increases, we have higher sync participation deviation. 



\subsubsection{Comparison with Other Synchronization Protocols}

We evaluate the performance of our algorithm (\textit{Fast\_Sync}) by comparing it with the BSP, SSP and DSSP synchronization protocols frequently used in training distributed machine learning models. For BSP, we fix the synchronization barrier at the time point where the last worker finishes executing the task before the sync point. Thus, fast workers need to wait for slow workers at the synchronization barrier. For SSP, we set the staleness threshold $s$ to $3$ and $5$ with each threshold unit being equivalent to $5ms$. For DSSP, we set $s$ to $3$ and the $r_{max} = 7$; this is the maximum allowable execution distance between the fastest and lowest worker beyond $s$. We consider a task graph with two asynchronous tasks, a single sync task and two local tasks. For each iteration, we split the execution time into computation, communication and clustering times.

\begin{figure*}[!ht]
\begin{multicols}{3}
  \centering
     \includegraphics[width=\linewidth]{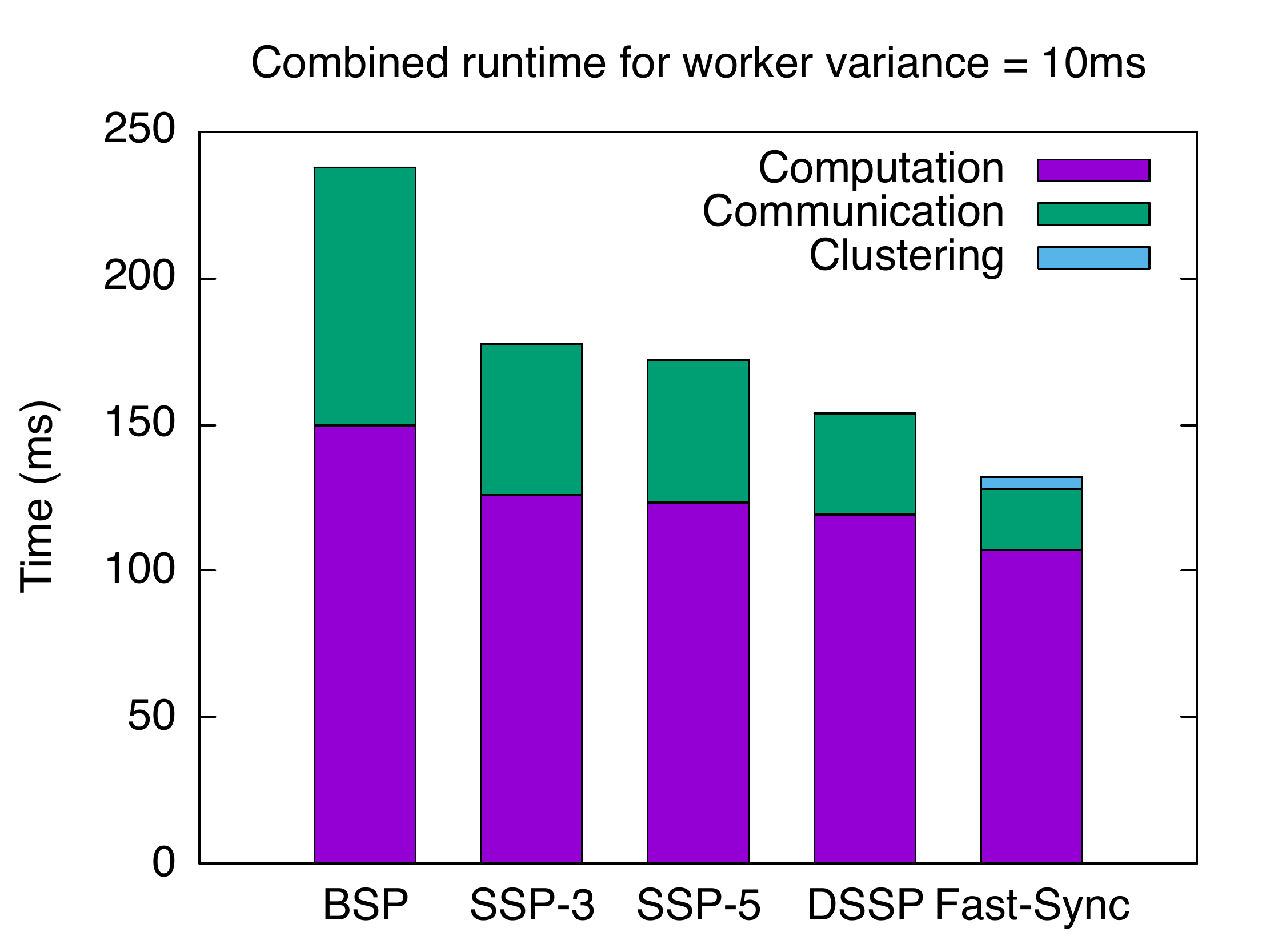}\par
    \caption{Average computation and communication times for worker execution variance of $10ms$ with $20$ workers.}
    \label{fig:combined_plot_var_10}
    \includegraphics[width=\linewidth]{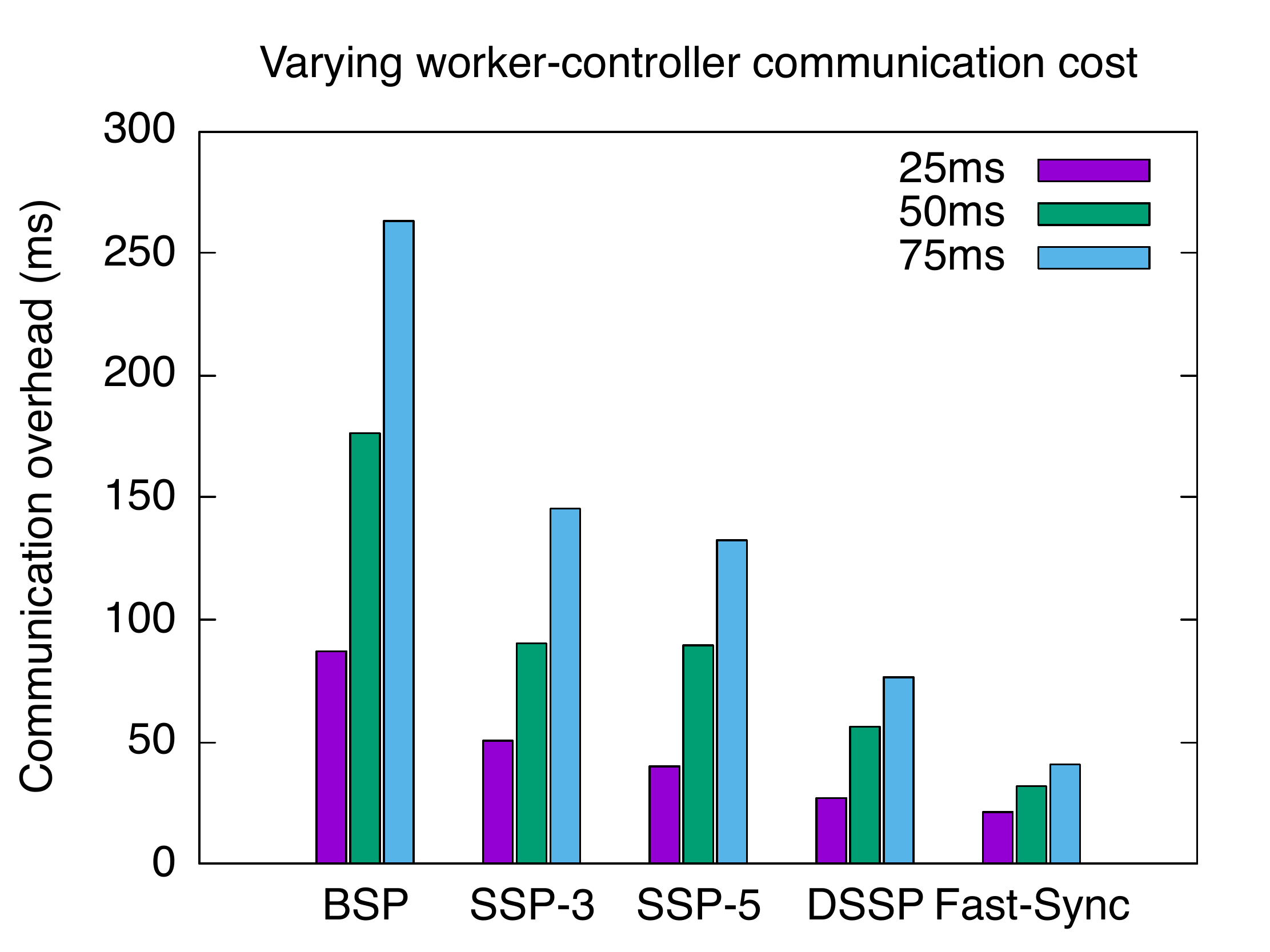}\par
    \caption{Average communication overhead for varying worker-controller message cost for $20$ workers.}
    \label{fig:combined_comm}
    \includegraphics[width=\linewidth]{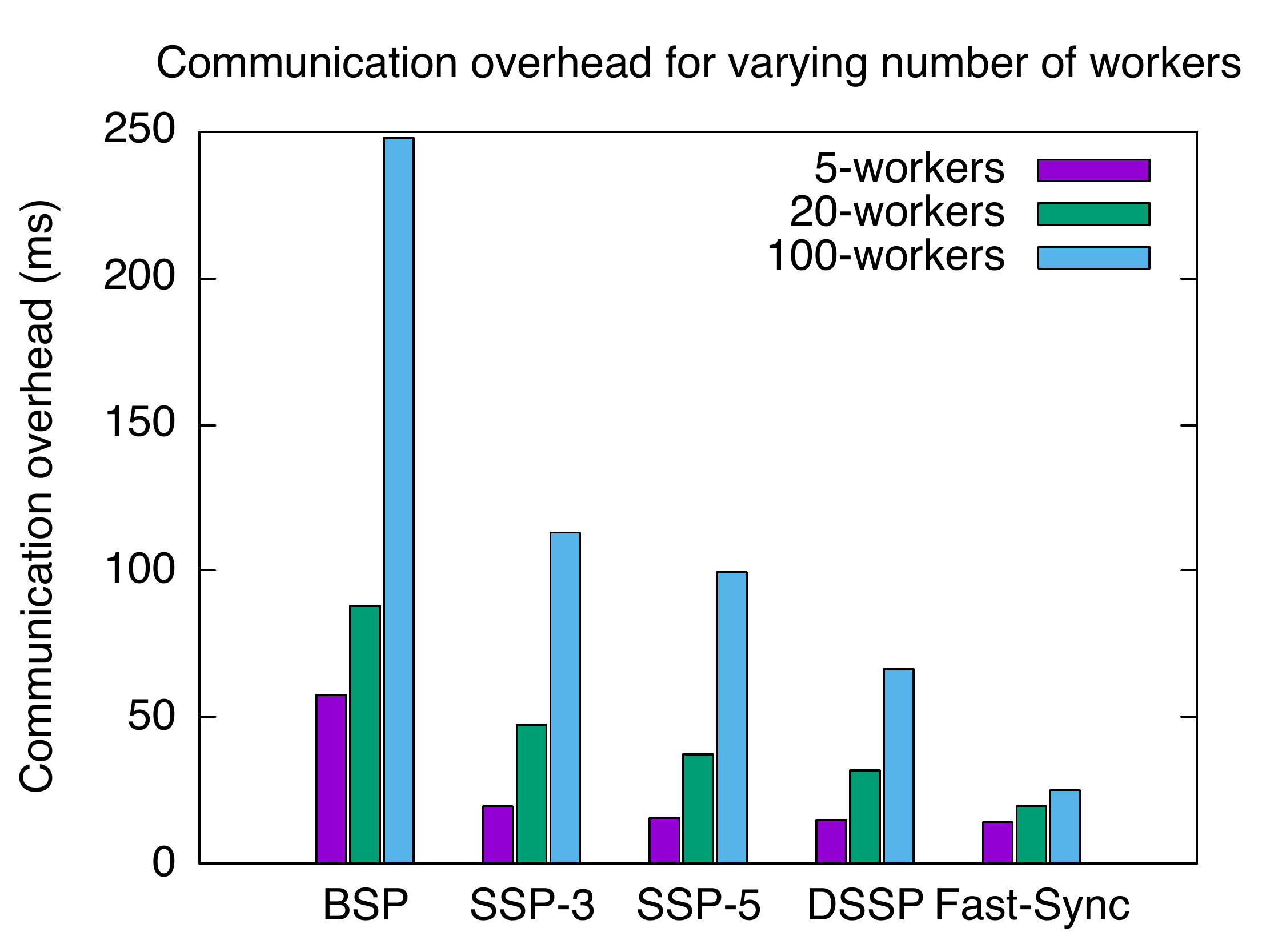}
    \caption{Average communication overhead for varying number of workers.}
    \label{fig:combined_scale}

\end{multicols}
\end{figure*}

To measure the effect of worker heterogeneity on the algorithms, we measure the execution runtimes for our algorithm and the other synchronization models with worker execution time variance of $2ms$ and $10ms$ across $20$ workers as shown in Fig.~\ref{fig:combined_plot_var_2} and~\ref{fig:combined_plot_var_10} respectively. Fig.~\ref{fig:combined_plot_var_2} shows the average runtime for our algorithm ($Fast\_Sync$), BSP, SSP and DSSP with worker execution time variance of $2ms$ while Fig.~\ref{fig:combined_plot_var_10} shows for varying worker execution variance of $10ms$. Our algorithm performs best in both cases, both in terms of computation time and communication cost. DSSP performs almost as good as our algorithm with BSP performing worst in both cases. Our algorithm outperforms DSSP due to the fact that DSSP is heavily reliant on the assumption that worker execution times does not vary (or varies minimally) over different iterations. 

We vary the worker-controller communication cost from $25ms$ to $75ms$ to measure the impact of network on all algorithms as shown in Fig.~\ref{fig:combined_comm}. Our algorithm incurs the least communication overhead for any given worker-controller communication cost. DSSP performs better than both SSP's while BSP performs worst. The communication overhead incurred by our algorithm increases at a lower rate compared to the other algorithms as the worker-controller communication cost is increased. This is because our algorithm has a bounded number of messages sent within the system.

We increase the number of workers from $5$ to $100$ to measure how scalable the algorithms are in terms of average communication overhead. Our algorithm outperforms other algorithms for increasing number of workers as shown in Fig.~\ref{fig:combined_scale}. Increasing the number of workers has a smaller impact on our algorithm compared to the other algorithms where the communication overhead is directly proportional to the number of workers. All of the other algorithms have a significant increase in the communication overhead as the number of workers is increased from $5$ to $100$.

The BSP, SSP and DSSP algorithms have been proven to converge. The BSP algorithm will always converge but the runtime is heavily affected stragglers. The SSP and DSSP algorithms will converge provided the staleness threshold is within some bound. Our scheme ensures that we have at least a certain ratio of devices to the available before synchronization proceeds. This helps in ensuring that the algorithm will converge.

\begin{figure*}[!ht]
\begin{multicols}{3}
  \centering
     \includegraphics[width=\linewidth]{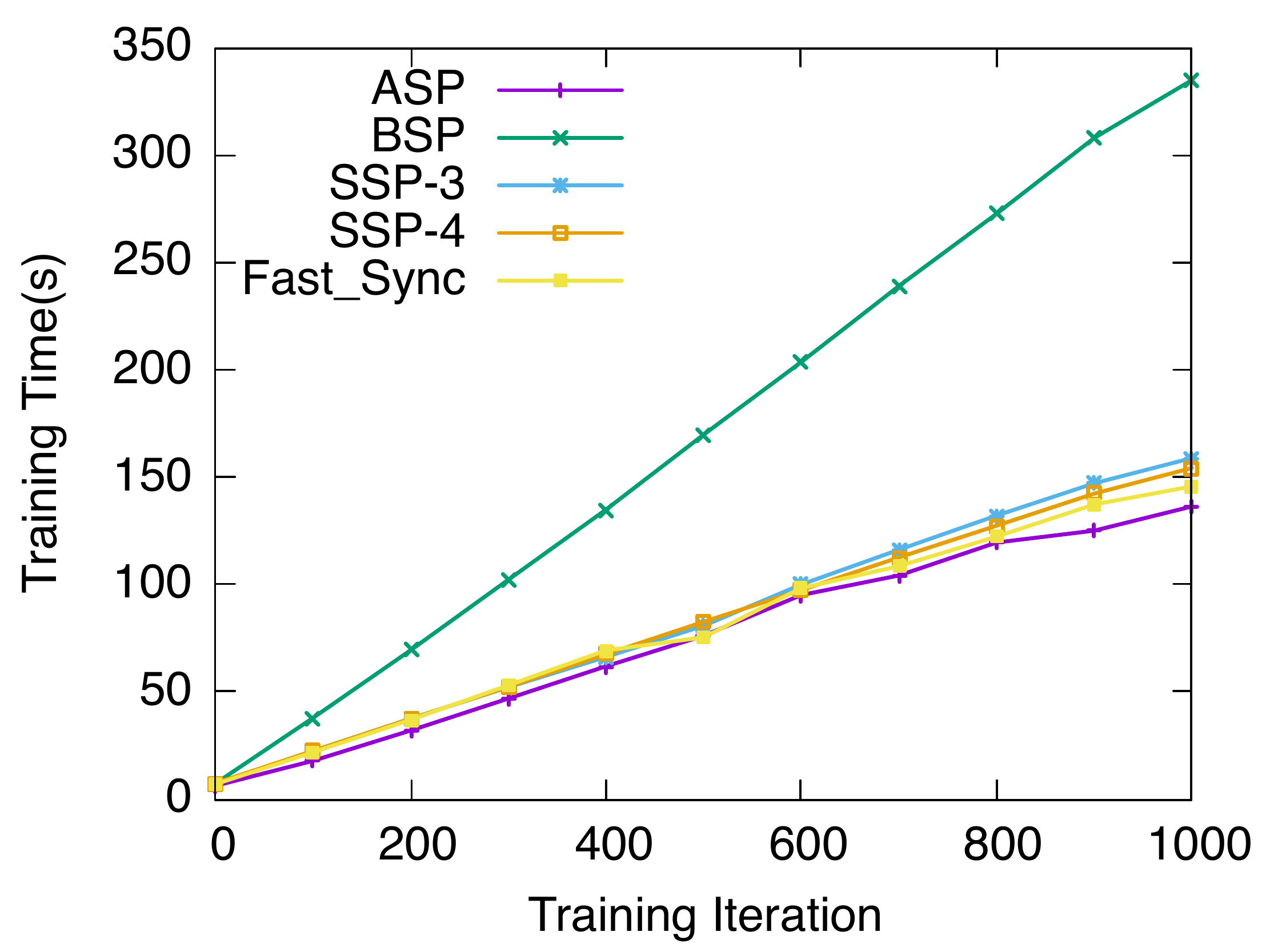}\par
    \caption{Training time versus training iteration for $5$ workers.}
    \label{fig:training_5}
    \includegraphics[width=\linewidth]{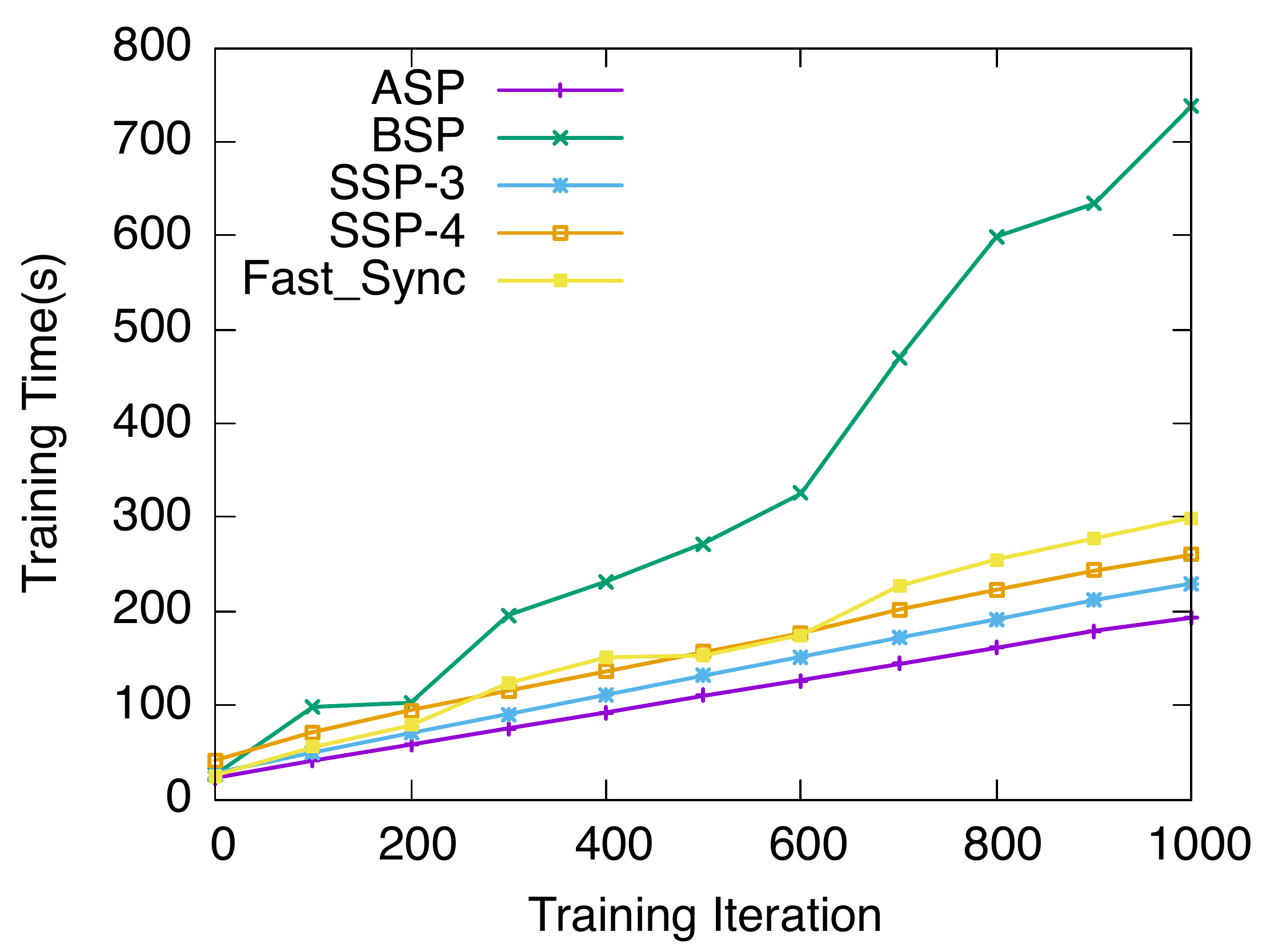}\par
    \caption{Training time versus training iteration for $20$ workers.}
    \label{fig:training_20}
    \includegraphics[width=\linewidth]{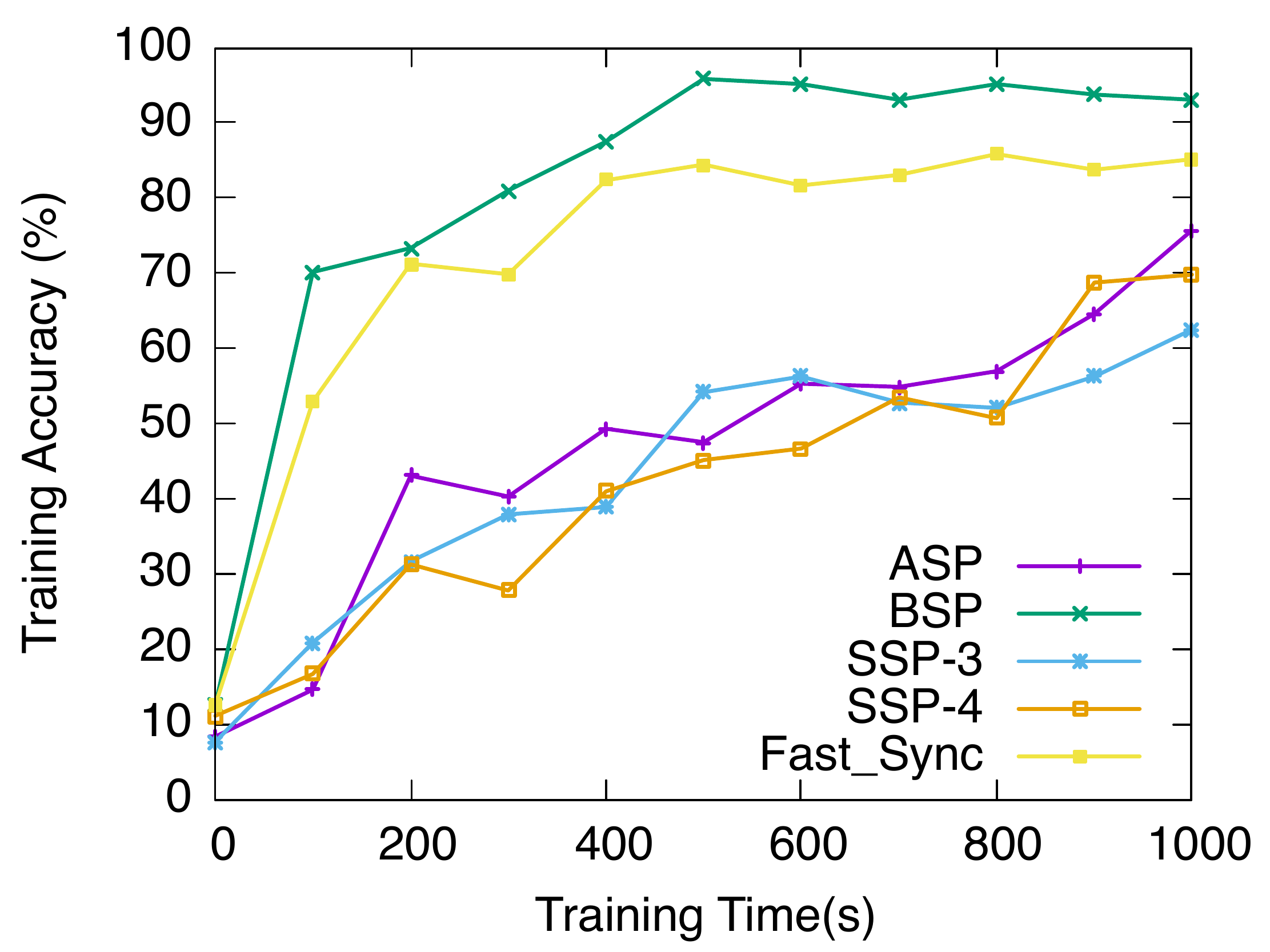}
    \caption{Training accuracy versus training time for $20$ workers.}
    \label{fig:accuracy20}

\end{multicols}
\end{figure*}

\subsection{Simulation Validation}

A lot of studies have been conducted comparing the BSP, SSP and DSSP synchronization schemes for parameter server models in distributed machine learning and neural networks. The results reported in ~\cite{ho2013more} and ~\cite{cipar2013solving} show that SSP converges to a consensus faster than BSP. The time to reach convergence for SSP reduces as the number of machines increase unlike in BSP. Both the BSP and  SSP have been found to converge provided the staleness threshold for the SSP is within some bound. The results in~\cite{zhao2019dynamic} show that DSSP converges faster than SSP in the same corresponding staleness threshold range.

Our results are similar to those in ~\cite{ho2013more}, ~\cite{cipar2013solving} and~\cite{zhao2019dynamic} with regards to running time of the training process. BSP takes a longer time to complete a specified number of iterations compared to SSP and DSSP. DSSP takes the shortest amount of time. However, our algorithm outperforms BSP, SSP and DSSP in terms of running time to complete a number of training iterations.

\subsection{Implementation}
 
We evaluate the performance of our algorithm compared to three other parameter server synchronization models (ASP, BSP and SSP). We aim to find how our algorithm compares to the other models in terms of accuracy and training time. We implement all the parameter server synchronization models in Ray. We train a simple 2D convolution neural network model with a batch size of $16$ on the KMNIST dataset\footnote{https://pytorch.org/docs/stable/torchvision/datasets.html\#kmnist} consisting of $70,000$ $28x28$ gray scale images ($60,000$ training and $10,000$ testing set examples). Each model is trained on a machine with $2.4$GHz Intel Core i$5$ machine with $12$GB of memory dedicated to the workers and $3$GB dedicated to the parameter server.

We compare the training accuracy and training time of our algorithm, ASP, BSP and SSP models for different number of workers. Fig~\ref{fig:training_5} and~\ref{fig:training_20} shows the training time versus training iterations for our algorithm (\textit{Fast\_Sync}), ASP, BSP, SSP-3 (staleness threshold $= 3$ iterations) and SSP-4 (staleness threshold $= 4$ iterations) for $5$ and $20$ workers respectively. ASP reaches $1000$ training iterations fastest closely followed by SSP, then our algorithm for both $5$ and $20$ workers. BSP takes the longest time to reach $1000$ training iterations as expected. However, BSP reaches a higher accuracy in a shorter amount of time $70\%$ in $100$s followed by our algorithm $54\%$ in $100$s. Our algorithm performs better than the ASP and SSP models in terms of accuracy for the same training time.

\section{Conclusion}

In this paper, we present a game-theoretic synchronization approach for AI application tasks. Our approach reduces the number of messages needed in reaching synchronization through the use of clustering and a late notification protocol. Existing protocols such as BSP, SSP and DSSP decide the synchronization time only when the workers get to the synchronization point. A lot of messages are then needed in reaching a consensus on synchronization. We develop a game to help in deciding the optimal number of synchronization options and determining their parameters. Thus, during runtime, workers do not need to communicate with each other to reach, postpone or abort synchronization. The only messages sent during the synchronization process are bounded late notifications.

We report on a simulation study that evaluates the benefits of our fast synchronization scheme. In particular, we explore the performance of our synchronization scheme under different operation conditions. We show that our scheme performs well with increasing number of workers and increasing heterogeneity among workers. We compare our scheme with BSP, SSP (with different staleness thresholds) and DSSP and show that our scheme performs better or as well as BSP, SSP and DSSP.

We implement our synchronization scheme and compare it with other parameter server synchronization schemes (ASP, BSP and SSP) by training a 2D convolution neural work using the KMNIST dataset. Our algorithm performs better than ASP and SSP (with staleness threshold of 3 and 4 iterations respectively) in terms of training accuracy for the same training duration. 

One area of future work is to extend the current synchronization game to more than two clusters. Another is to fully implement our synchronization scheme into a framework and programming language for AI application tasks. This would allow us to evaluate our synchronization scheme under real-life scenarios.

\bibliographystyle{IEEEtran}
\bibliography{main}

\begin{IEEEbiography}[{\includegraphics[width=1in,height=1.25in,clip,keepaspectratio]{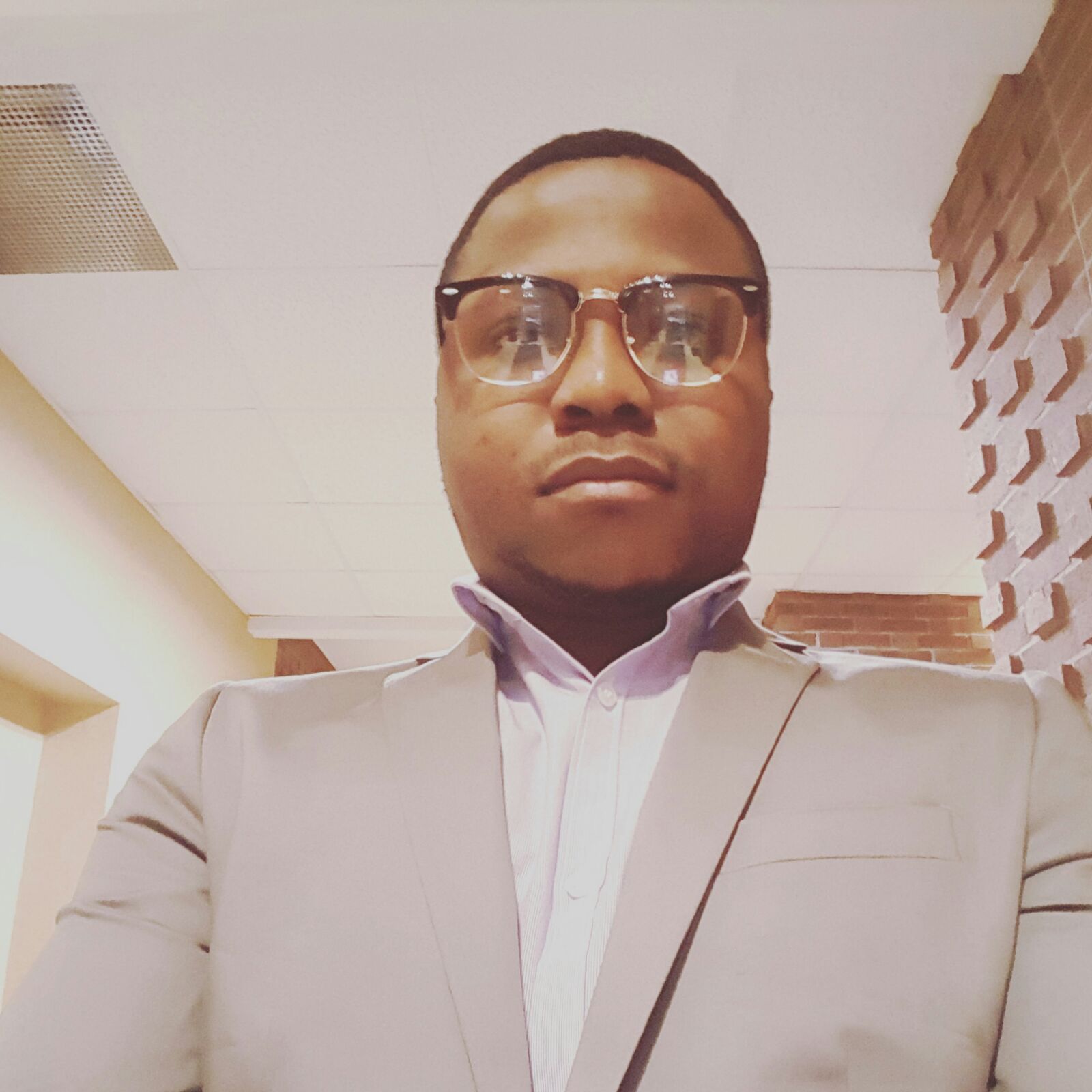}}]{Richard Olaniyan}
 is a PhD student in the School of Computer Science, McGill University, Montreal, Canada being sponsored by the Presidential Scholarship Scheme of the Nigerian Government/Petroleum Technology Development Fund (PTDF) Nigeria. He is currently doing an internship with Ericsson Canada. He received his MSc degree in Computer Science at the University of Edinburgh, United Kingdom in 2015. He Received his BSc degree in Computer Engineering from Obafemi Awolowo University, Ile-Ife, Nigeria in 2011, where he graduated as the best student in the department bagging two awards. His research interests include synchronization and scheduling in clouds, clusters, fog computing, edge computing, vehicular clouds and computing models. 
\end{IEEEbiography}

\begin{IEEEbiography}[{\includegraphics[width=1in,height=1.25in,clip,keepaspectratio]{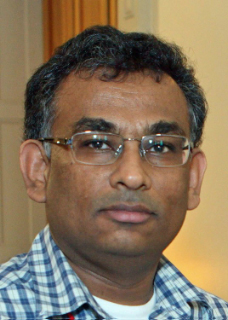}}]{Muthucumaru Maheswaran}
 is an associate professor in the School of Computer Science at McGill University. He got a PhD in Electrical and Computer Engineering from Purdue University, West Lafayette and a BScEng degree in Electrical and Electronic Engineering from the University of Peradeniya, Sri Lanka. He has researched various issues in scheduling, trust management, and scalable resource discovery mechanisms in Clouds and Grids. Many papers he co-authored in resource management systems have been highly cited by other researchers in the area. Recently, his research has focused in security, resource management, and programming frameworks for Cloud of Things. He has supervised the completion of 8 PhD theses in the above areas. He has published more than 120 technical papers in major journal, conferences, and workshops. He holds a US patent in wide-area content routing.
\end{IEEEbiography}
\vfill
\end{document}